\documentclass[onecolumn,10pt]{article}
\usepackage[top=.75in, bottom=.75in, left=.75in, right=.75in]{geometry}
\usepackage{amsmath,amsfonts,amscd,amssymb}
\usepackage{graphicx}
\usepackage{epstopdf}
\usepackage{overpic}
\usepackage{cancel}
\usepackage{rotating}
\usepackage{url}
\usepackage{caption}
\usepackage{color}
\usepackage{rotating}
\usepackage{multirow}
\usepackage{wrapfig}
\usepackage{mathtools}
\usepackage{subeqnarray}
\usepackage{setspace}
\usepackage{palatino} 
\usepackage[numbers,sort&compress]{natbib}

\usepackage[bottom,flushmargin,hang,multiple]{footmisc}
\usepackage{lipsum}
\newcommand\blfootnote[1]{%
  \begingroup
  \renewcommand\thefootnote{}\footnote{#1}%
  \addtocounter{footnote}{-1}%
  \endgroup
}

\definecolor{header1}{cmyk}{0,0,0,1}

\usepackage[utf8]{inputenc}

\usepackage[normalem]{ulem}
\usepackage{color}

\title{\vspace{-.125in}{\huge\selectfont \textbf{Machine Learning for Partial Differential Equations}}\vspace{-.075in}}

\author{\normalsize{Steven L. Brunton$^{1*}$ and J. Nathan Kutz$^{2}$}\\
\footnotesize{$^1$ Department of Mechanical Engineering, University of Washington, Seattle, WA 98195, United States}\\
\footnotesize{$^2$ Department of Applied Mathematics, University of Washington, Seattle, WA 98195, United States \vspace{-.2in}}
}

\date{}

\begin{document}
\maketitle

\blfootnote{$^*$ Corresponding author: sbrunton@uw.edu}
\vspace{-.2in}
\begin{abstract}
Partial differential equations (PDEs) are among the most universal and parsimonious descriptions of natural physical laws, capturing a rich variety of phenomenology and multi-scale physics in a compact and symbolic representation.  This review will examine several promising avenues of PDE research that are being advanced by machine learning, including: 1) the discovery of new governing PDEs and coarse-grained approximations for complex natural and engineered systems, 2) learning effective coordinate systems and reduced-order models to make PDEs more amenable to analysis, and 3) representing solution operators and improving traditional numerical algorithms.  
In each of these fields, we summarize key advances, ongoing challenges, and opportunities for further development.  
\end{abstract}

\section{Introduction}\label{Sec:Intro}

Partial differential equations (PDEs) have been a cornerstone of mathematical physics and engineering design for over 250 years, since the introduction of the one-dimensional wave equation by d’Alembert in 1752~\cite{brezis1998partial}. 
PDEs provide a formal mathematical infrastructure for relating how quantities
of interest change in several variables, typically space and time.  As such, PDEs provide a foundational description of the governing equations of many canonical spatio-temporal physical systems, including electrodynamics, quantum mechanics, fluid mechanics, heat transfer, etc.   
Today, nearly every aspect of our engineered world is based in some way on the predictive capability of PDEs, from structural modeling of buildings and bridges, to the design of aircraft and other vehicles, to the thermal and electromagnetic management systems in modern portable electronics.  
In general, we will consider a PDE for $u(x,t)$
\begin{equation}
  u_t + N (u, u_x , u_{xx}, \cdots, x, t; \mu) = f(x)
  \label{eq:PDE}
\end{equation}
on a spatial domain $x\in[0,L]$,for time $t\in[0,T]$, and where the parameter $\mu$ denotes a parametric dependency. The initial and boundary conditions are given by
\begin{subequations}\label{eq:ICBC}
\begin{align}
\text{IC:}&\quad  u(x,0)=u_0(x) \\
\text{BC:}&\quad  \alpha_1 u (0,t) +\beta_1 u_x(0,t) = g_1(t) \,\, \text{and} \,\, \alpha_2 u(L,t) + \beta_2 u_x(L,t) = g_2(t).
\end{align}
\end{subequations}
This may be generalized to systems of several spatial variables, or a system with no time dependence.  

In the past half-century, the advent of computing has produced two revolutions in our capability to analyze and solve PDEs.
In the first, closed-form analytic solution techniques, which typically rely on linearity and superposition principles, have given way to diverse computational approximations based upon finite difference, finite element, and spectral techniques. 
These computational approaches significantly expand the complexity of behaviors and solutions that can be analyzed.  Importantly, scientific computing has allowed for the study of nonlinear systems for which our analytic techniques typically fail. Additionally, complex boundary conditions, difficult geometries, and multiscale interactions can all be characterized within this framework. 
Thus, the combination of analytic and computational methods for solving PDEs has driven critical technological advancements in many industries since the 1960s.
However, modern PDE systems are often nonlinear, complex, and high-dimensional, rendering analytic techniques ineffective and computational methods intractable. 
More recently, the ongoing machine learning revolution is providing an entirely new approach for solving PDEs, based on the increasing wealth of high-quality data generated from both simulations and experiments.  Indeed, the emergence of machine learning methods in the last decade have allowed the community a significantly different approach to modeling the dynamics of PDEs, allowing for the learning and construction of proxy, reduced-order models which are faithful representations of the full, high-dimensional complex dynamics.
Although machine learning has been applied to study PDEs for nearly three decades~\cite{dissanayake1994neural}, several key advances in computational capabilities and algorithms are dramatically accelerating these efforts in the past decade. 

In this review, we explore several avenues of PDE research that are being advanced by machine learning: 
\begin{itemize}
    \item \textbf{Governing equations and coarse-grained closures:} Emerging techniques in symbolic regression and new high-fidelity measurements are making it possible to learn new PDEs for systems that are not amenable to first-principles and by-hand derivations.  Systems in neuroscience and epidemiology, as well as systems from traditional physics, such as plasma dynamics, non-Newtonian fluids, and active matter, are all candidates for improved PDE descriptions.  Moreover, there are many systems where we have accurate governing equations, but they are too computationally expensive to resolve at all scales of the physics.  Thus, one must resort to coarse-grained PDEs.  Machine learning is enabling tremendous progress in this traditionally challenging field, for example in turbulence modeling and the modeling of geophysical fluids.  Several other fields stand to benefit, including material science and biology.  
    \item \textbf{Coordinate systems and reduced representations:} Solution techniques for PDEs are intimately tied to a coordinate system.  For example, the Fourier transform is the coordinate system that diagonalized Laplace’s equation.  However, for nonlinear PDEs there is generically no coordinate system that simplify the equations.  Advances in modern Koopman operator theory are providing a powerful new perspective for finding effective coordinates even for nonlinear systems.  Similarly, reduced-order models provide a reduction of a PDE to a much simpler ODE system that is tailored to the specific configuration and parameters of interest.  Machine learning has rapidly been adopted as a new technique for ROMs, as it shares a significant overlap and history with this field of applied mathematics.  For many iterative design optimization and control applications, ROMs are critical, as there is a tradeoff between the accuracy of a solution and the cost of computing it.  
    \item \textbf{Numerical solutions and operator learning:} Another major avenue of research is focused on learning the solution operator of complex PDEs, trained from limited amounts of high-fidelity data.  These approaches are quite flexible and offer many advantages, including the ability to re-mesh solutions flexibly.  In related work, researchers are currently using machine learning to improve traditional scientific computing workflows, for example to improve pre-conditioning and to learn improved stencils for shocks and discontinuities.
\end{itemize}

Our goal with this review is to provide a brief summary and organization of the rapid progress in this field, along with a high-level perspective on the ongoing challenges and avenues of future opportunity.  Although machine learning is changing how we learn, represent, and solve PDEs, many things haven't changed.  We still seek interpretable and generalizable representations of the governing equations and their solutions.  We still use techniques from scientific computing to integrate many of these models and propagate their uncertainty.  And we still use the same iterative design optimization and control algorithms, now wrapped around machine learned models and solutions. 
The importance of embedding physics into machine learning has also become increasingly clear in recent years~\cite{Loiseau2017jfm,cranmer2020lagrangian,Brunton2020arfm,wang2020incorporating,wang2020towards,brandstetter2022clifford,de2020gauge,brandstetter2022lie,brandstetter2022message,karniadakis2021physics,Brunton2022book}.  
Incorporating physics into the learning process makes it possible to achieve more accurate solutions, with more compact architectures, and from less and noisier training data.  

\section{Learning Governing Equations and Coarse-Grained Closures}\label{Sec:Sec1}
Despite the significant progress over the past half century in the computational solution to PDEs, the fundamental process of how we \emph{derive} PDEs has remained largely unchanged since the 1700s: equations are typically derived from governing conservations laws and symmetries using control volume techniques, with the pill box derivation remaining the standard in university classrooms.  
The current availability of vast quantities of measurement data from both familiar and exotic new fields of science and engineering are providing entirely new opportunities to \emph{learn} PDEs, and thus perhaps underlying laws, principles, invariances and symmetries. 
This is one of the most promising avenues of modern scientific discovery, as we are on the cusp of the automatic and data-driven discovery of entirely new physics for systems that have alluded researchers for decades, if not centuries.  
Moreover, learning PDEs from data has several advantages over alternative approaches of using deep learning to ``mimic” the behavior of a complex system.  
First, PDEs are inherently interpretable, in the sense that they can be tied directly to geometry, conservation laws, symmetries, constraints, etc.
Second, PDEs are highly generalizable, in that by changing the boundary conditions and parameters, a wide variety of phenomena and bifurcations may emerge. 

Machine learning is ideal for representing arbitrarily complex input--output relationships from data.  However, many techniques result in opaque models that are not interpretable.  In contrast, symbolic regression techniques~\cite{Bongard2007pnas,Schmidt2009science,Brunton2016pnas} is a class of machine learning that results in highly interpretable symbolic models by design.  Recently, the sparse identification of nonlinear dynamics (SINDy) algorithm~\cite{Brunton2016pnas} has been extended to learn partial differential equations from data~\cite{Rudy2017sciadv,Schaeffer2017prsa}. 
The resulting algorithm, called PDE-FIND, provides new opportunities for scientific discovery by enabling the learning of new PDEs for unknown physics as well as coarse-grained closure models.  

\begin{figure}
    \centering    \includegraphics[width=\textwidth]{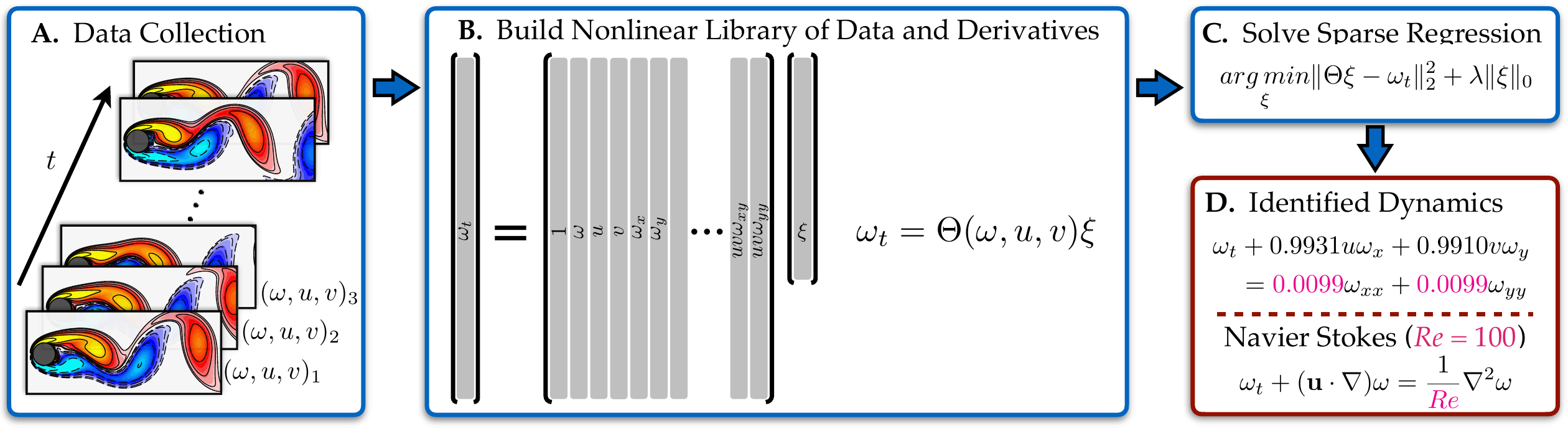}
    \caption{Sparse regression procedure to discover PDEs from data, demonstrated on the Navier--Stokes equations. {A.}~Data is collected as snapshots\index{snapshot} of a solution to a PDE. {B.}~Numerical derivatives are taken and data is compiled into a large matrix $\mathbf{\Theta}$, incorporating candidate terms for the PDE. {C.}~Sparse regression is used to identify active terms in the PDE. {D.}~Active terms in $\boldsymbol{\xi}$ are synthesized into a PDE. Modified from Rudy et al. \cite{Rudy2017sciadv}.}
    \label{fig:PDEFIND}
\end{figure}

The basic idea behind SINDy and PDE-FIND is to approximate the time derivative of the state of a dynamical system as a sparse linear combination of candidate terms that can describe the dynamics:
\begin{align}\label{eq:PDEFIND}
    u_t \approx \Theta(u,u_x,u_{xx},uu_x,\cdots)\xi
\end{align}
where $\Theta$ is the library of candidate terms and $\xi$ is a sparse vector that selects the relevant terms needed to describe the dynamics.  There are several algorithms to learn these sparse dynamics, typically based on sparse optimization.  
Figure~\ref{fig:PDEFIND} illustrates this approach to learn the Navier-Stokes equations for a fluid flow. 

In a relatively short time, PDE-FIND has been used to rediscover several models from classical physics, as well as to discover entirely new models of modern interest.  
Rediscovery, or the process of recapitulating existing theories with modern techniques, is a reasonable first step when testing out a new method.  If the algorithm doesn't work on problems where we know the answer, it is unreasonable to expect it to yield new insights for more challenging systems.  In addition, testing an algorithm on a known system may provide considerable insights.  In early papers, PDE-FIND was applied to rediscover several canonical PDEs from physics, including the Navier-Stokes equation for fluid flows, and the Schrodinger equation for quantum mechanics.  

\begin{figure}
    \centering    
    \includegraphics[width=.95\textwidth]{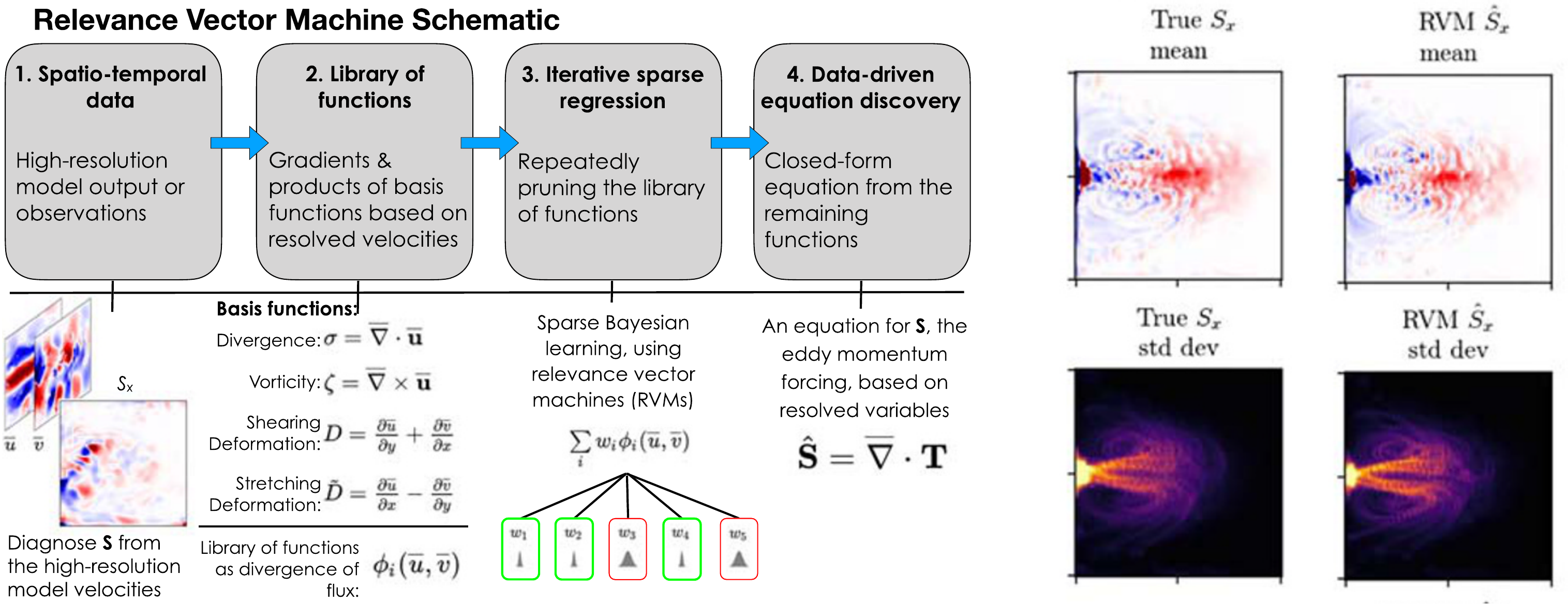}
    \caption{Illustration of sparse Bayesian PDE discovery applied to LES closure modeling in large-scale geophysical fluid simulations.  Modified from Zanna and Bolton~\cite{zanna2020data}.}
    \label{fig:Zanna}
\end{figure}

More recently, PDE-FIND is being applied to entirely new fields of physics and natural sciences with promising results.  Many of these advances have come in the field of fluid dynamics~\cite{gurevich2019robust,zanna2020data,alves2020data,Reinbold2020pre,suri2020capturing,beetham2020formulating,beetham2021sparse,schmelzer2020discovery,reinbold2021robust}, where there are many open problems related to constituative modeling and turbulence closure modeling~\cite{pope1975more,Ling2016jfm,Duraisamy2019arfm,Brunton2020arfm,ahmed2021closures}.   PDE-FIND has become a powerful tool for closure modeling of fluids, enabling both Reynolds averaged Navier-Stokes (RANS) closures~\cite{beetham2020formulating,beetham2021sparse,schmelzer2020discovery} and large eddy simulation (LES) closures~\cite{zanna2020data}.  Figure~\ref{fig:Zanna} shows an approach based on a sparse Bayesian formulation of the problem that discovers LES closure models for large-scale atmospheric simulations that preserve underlying symmetries and structure~\cite{zanna2020data}.  In addition to closure modeling for known PDEs, such as the Navier-Stokes equations, there are also efforts to learn additional physics terms corresponding to currently unmodeled mechanisms, for example in plasma physics~\cite{alves2020data}, viscoelastic flows, granular materials, and active matter~\cite{supekar2023learning}.  And while the incompressible Navier-Stokes equations are a remarkably accurate model for incompressible fluid flows, the magnetohydrodynamics (MHD) equations are much more of an approximation to plasma physics.  Researchers are currently leveraging PDE-FIND to learn additional correction and closure terms for the MHD equations to more accurately match high-fidelity particle in cell simulations~\cite{alves2020data}. 

To model new physical and biological processes, several methodological innovations have been introduced into the PDE discovery framework.  
Perhaps the most fundamental advance is the introduction of the \emph{weak form} PDE-FIND optimization by Messenger and Bortz~\cite{messenger2021bweak,messenger2021weak}.  This approach solves the sparse regression problem after integrating the data over random control volumes, providing a dramatic improvement to the noise robustness of the algorithm.  Weak form optimization may be thought of as a generalization of the integral SINDy~\cite{Schaeffer2017pre} to PDE-FIND.  
Further improvements to noise robustness and limited data may be obtained through ensembling techniques~\cite{fasel2022ensemble}, which use robust bagging to learn inclusion probabilities for the sparse terms $\xi$ in Eq.~\eqref{eq:PDEFIND} from limited and noisy data.  
These methodological innovations, and more, have been assembled into the open-source PySINDy software library~\cite{kaptanoglu2021pysindy}, reducing the barrier to entry when applying these methods to new problems.  

In addition to PDE-FIND, additional techniques for learning PDEs from data include PDE-NET~\cite{long2018pde,long2019pde} and the Bayesian PDE discovery from data~\cite{atkinson2020bayesian}. An earlier approach by Schaeffer et al.~\cite{Schaeffer2013pnas} importantly recognized that many PDE solutions are sparse in a suitably transformed solution space, introducing one of the first notions of sparsity in the field of PDEs.  Other symbolic learning techniques are also promising, including symbolic learning on graph neural networks~\cite{cranmer2019learning,cranmer2020discovering,sanchez2020learning}.

In addition to discovering new PDEs and closure models, recent work by Callaham et al.~\cite{callaham2021learning} has shown that it is possible to cluster spatiotemporal data by which subset of the terms in a PDE are active in what regions of space and time.  The resulting algorithm uncovers different regimes where a subset of the physics is active and in a dominant balance, providing a data-driven clustering based on the active terms in the PDE.  In related work, Bakarji et al.~\cite{bakarji2020data} leveraged sparse model learning to automate the Buckingham Pi procedure for learning nondimensional quantities that mediate the transition across these dominant balance regimes and control bifurcations in system behavior.  For example, these methods were applied to study boundary layers, where the classic boundary layer regions and Blasius scaling laws were recapitulated.  

There are many exciting open problems in physical, engineered, and biological systems where PDE discovery may play an important role.  In addition to discovering new mechanistic models and closures for problems in fluids and plasmas, granular materials, non-Newtonian and active matter, there are significant opportunities to learn coarse-grained models of biological systems, such as collective dynamics of many biological agents, the dynamics of bacterial colonies, spatiotemporal models in neuroscience and organized biological matter such as muscles.   
New methodological innovations will likely be required for these systems, for example to model non-stationarity and non-local interactions, as well as spatial heterogeneity.  
However, there is a large and active community working on addressing these issues, and the quality and quantity of measurement data is increasing every day.  

\section{Learning Coordinates and Reduced Representations}\label{Sec:Sec2}
In the history of science, many breakthroughs in learning governing equations have been preceded by learning an effective coordinate system.  In recent decades, there are two dominant perspectives on effective coordinates related to PDEs, which we will explore here.  First, advances in Koopman operator theory~\cite{Koopman1931pnas,Mezic2004,Mezic2005nd,Mezic2013arfm,Rowley2009jfm,Kutz2016book,Budivsic2012chaos,Brunton2022siamreview} are making it possible to learn coordinate systems in which nonlinear dynamics appear linear.  Second, it is often possible to reduce the dimension of a high-dimensional spatiotemporal system through a coordinate transformation to obtain a reduced-order model, which balances accuracy and efficiency.  Both of these fields are rapidly progressing with advances in machine learning.

\subsection*{Linearizing coordinate transformations}
By construction, the Koopman operator~\cite{Koopman1931pnas} is a {\em linear}, infinite-dimensional operator that acts on the Hilbert space $\mathcal{H}$ of \emph{all} scalar measurement functions $g$.  The Koopman operator acts on functions of the state space of the dynamical system,  trading nonlinear finite-dimensional dynamics for linear infinite-dimensional dynamics. It can be further generalized to map infinite-dimensional nonlinear dynamics to infinite-dimensional linear dynamics by appropriate choice of observables.   The advantage of the Koopman representation is obvious:  the linear problem can be solved using a standard spectral decomposition (\ref{eq:eigenfunction}).   Thus the inversion of the operator is achieved by construction of the Koopman operator and by projecting into its eigenfunction space.
Cover's theorem~\cite{cover1965geometrical} represents a corresponding theory for how the projection to infinite-dimensions allows for {\em linear} separability of data, and thus the underlying success of kernel methods and support vector machines. 

The Koopman operator is defined for discrete-time dynamical systems.
A continuous dynamical system will induce a discrete-time dynamical system given by the flow map ${\bf F}_t:\mathcal{M}\rightarrow\mathcal{M}$, which maps the state ${\bf x}(t_0)$ to a future time ${\bf x}(t_0+t)$:
\begin{eqnarray}
{\bf F}_t({\bf x}(t_0)) = {\bf x}(t_0+t) = {\bf x}(t_0) + \int_{t_0}^{t_0+t}\bf{f}({\bf x}(\tau))\,d\tau\,.
\end{eqnarray}
This induces the discrete-time dynamical system
\begin{eqnarray}
{\bf x}_{k+1} = {\bf F}_t({\bf x}_k),\label{Eq:Dynamics}
\end{eqnarray}
where ${\bf x}_k={\bf x}(kt)$.  The analogous discrete-time Koopman operator is given by $\mathcal{K}_t$ such that $\mathcal{K}_tg = g\circ{\bf F}_t$.  
Thus, the Koopman operator sets up a discrete-time dynamical system on the observable function $g$:
\begin{eqnarray}
\mathcal{K}_tg({\bf x}_k) = g({\bf F}_t({\bf x}_k)) = g({\bf x}_{k+1}).
\end{eqnarray}
The Koopman operator can be constructed using deep learning methods in order to enforce the above constraint on observables.  The result is a spectral decomposition capable of representing the dynamical solutions of interest.  Specifically, the eigenfunctions and eigenvalues of the Koopman operator ${\cal K}$ give a complete characterization of the dynamics ${\cal K} \varphi_k = \lambda_k \varphi_k$.  The functions $\varphi_k({\bf x})$ are Koopman eigenfunctions, and they define a set of intrinsic measurement coordinates, on which it is possible to advance these measurements with a \emph{linear} dynamical system.  A reduced-order linear model can be constructed by retaining the dominant Koopman eigenfunctions $\varphi_k$.

Such linearizing transforms for PDEs are not new.   Indeed, the Cole-Hopf transformation~\cite{hopf50,cole51} for solving diffusively regularized Burgers' equation was the first successful demonstration of such a technique, and the {\em Inverse Scattering Transform} (IST)~\cite{ablowitz1974inverse,ablowitz1981solitons}  generalized this framework for a class of completely integrable PDEs.  But recent data-driven modeling paradigms have given Koopman theory a modern interpretation in terms of dynamical systems theory~\cite{Mezic2004,Mezic2005nd}.  And even more recently,    
deep learning approaches have provided Koopman embeddings for dynamics using neural networks~\cite{lusch2018deep,Wehmeyer2017arxiv,Mardt2017arxiv,Takeishi2017nips,Yeung2017arxiv,Otto2017arxiv,Li2017chaos}.  This is in addition to enriching the observables of the {\em dynamic mode decomposition}~\cite{noe2013variational,nuske2014variational,Williams2015jnls,Williams2014arxivA,klus2017data,kutzPDE,page2018koopman}.  Importantly, Koopman theory attempts to approximate the dynamics with a linear operator while the work of Lu et al~\cite{lu2019deeponet,lu2021learning} and  Kovachki et al~\cite{kovachki2021neural} directly construct a nonlinear operator using neural networks.

\begin{figure}[t]
\centering
\includegraphics[width=.8\textwidth]{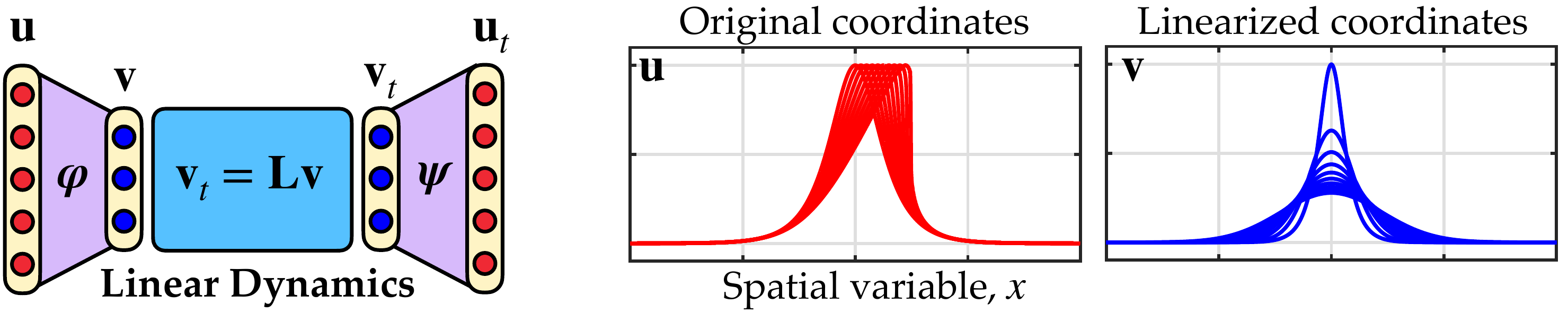}
 \caption{A schematic of a neural network architecture used to discover a Koopman linearizing coordinate transformation.  In this example, the nonlinear Burgers' equation is transformed into the linear heat equation.}
 \label{fig:NetworkArch}
\end{figure}

As an example deep learning architecture (see Fig.~\ref{fig:NetworkArch}), consider a neural network $f(\mathbf{u})$ that advances the state variable forward in time $\mathbf{u}_{k+1} = f(\mathbf{u}_k)$, and can be expressed by the formula
\begin{equation}
 f(\mathbf{u}) =  \varphi_d(\mathbf{K}( \varphi_e(\mathbf{u}))).
\end{equation}
The input of the the network $\mathbf{u}_k$ is the state vector at time $t_k$ and the output is the state vector at time $t_{k+1}$. The network consists of three parts: (i) the encoder $\varphi_e$, (ii) the linear dynamics $\mathbf{K}$, and (iii) the decoder $\varphi_d$. In this example, both the encoder and decoder are split into two parts where the outer encoder is $\chi + \mathbf{I}$ and the inner encoder $\psi_e$ where
$ \varphi_e(\mathbf{u}) = \psi_e((\chi + \mathbf{I})(\mathbf{u}))$.  The outer encoder performs a coordinate transformation into a space in which the dynamics are linear. The inner encoder either (i) diagonalizes the system, (ii) reduces the dimensionality, or (iii) both. The decoder consists of the inner decoder $\psi_d$ and the outer decoder $\mathbf{\zeta} + \mathbf{I}$.  The inner and outer decoder are approximate inverses of the inner and outer encoder, respectively.  The loss function used to train the network accounts for the autoencoder loss, the outer autoencoder loss, the inner autoencoder loss, the prediction loss, the linearity loss and regularization of the weights~\cite{gin2021deep}.  The diversity of methods used to build a Koopman representation~\cite{lusch2018deep,Wehmeyer2017arxiv,Mardt2017arxiv,Takeishi2017nips,Yeung2017arxiv,Otto2017arxiv,Li2017chaos,noe2013variational,nuske2014variational,Williams2015jnls,Williams2014arxivA,klus2017data,kutzPDE,page2018koopman,gin2021deep} highlights the significant efforts by the community to leverage the eigenfunction expansion of the Koopman operator in order to construct the inverse operator. Koopman theory has also been recently combined with recurrent neural networks to predict turbulence time series~\cite{eivazi2021recurrent}.  

\subsection*{Reduced-order models}
There is a rich history of constructing reduced-order models for PDEs~\cite{Lumley:1970,Berkooz1993arfm,Holmes2012book,Taira2017aiaa,Taira2020aiaaj,Benner2015siamreview,qian2020lift,peherstorfer2016data,benner2020operator,peherstorfer2020stability} by representing the dynamics in a lower-dimensional subspace or submanifold.  
The bulk of these methods are based on classical dynamical systems theory~\cite{guckenheimer_holmes} and symmetry reduction and manifold arguments~\cite{mars_rowl,MarsdenMTAA,Marsden:MS}. 
The most common and classical approach to reduced-order modeling of fluids involves identifying a low-dimensional subspace, typically through proper orthogonal decomposition (POD), and then projecting the governing equations onto this subspace through Galerkin projection.  In this approach, the field variable $u(x,t)$ is approximated in a finite Galerkin expansion 
\begin{align}\label{eq:POD}
    u(x,t) \approx \bar{u}(x) + \sum_{k=1}^r a_k(t) \varphi_k(x)
\end{align}
where $\bar{u}(x)$ is the average field, $\varphi_k(x)$ are spatial POD modes and $a_k(t)$ are the amplitudes of these modes in time.  
This space-time separation of variables is usually computed through a singular value decomposition (SVD)~\cite{Brunton2022book}, and represents a data-driven generalization of the Fourier transform.  
Galerkin projection of the governing equations \eqref{eq:PDE} onto the orthogonal basis in \eqref{eq:POD} then yields an ordinary differential equation in the POD mode amplitudes $\mathbf{a}(t)$:
\begin{align}\label{eq:ROM}
    \frac{d}{dt}\mathbf{a}(t) = \mathbf{f}(\mathbf{a}(t)).
\end{align}
The resulting ODE is typically much more tractable to simulate rapidly for iterative design optimization and real-time feedback control, but it comes at the cost of only approximately capturing the dynamics of the system.  There are a number of well-known issues, such as stability issues of the reduced ODE, and limitations of this approach when applied to turbulence or other multiscale phenomena.  

There are several major advances to this reduced-order modeling pipeline with the advent of machine learning.  
First, it is possible to dramatically improve the dimensionality reduction in Eq.~\eqref{eq:POD} by replacing the linear subspace approximation with a nonlinear manifold approximation using a deep neural network autoencoder~\cite{Brunton2020arfm}.  This is the approach taken by Lee and Carlberg~\cite{lee2020model}. 
It is also possible to learn the dynamics in Eq.~\eqref{eq:ROM} by data-driven regression, instead of projection based methods, for example using dynamic mode decomposition (DMD) for linear models~\cite{Schmid2010jfm,Rowley2009jfm,Kutz2016book}, SINDy for nonlinear models~\cite{Loiseau2017jfm,Loiseau2018jfm,Deng2020JFM,Loiseau2020tcfd,guan2020sparse,kaptanoglu2020physics,kaptanoglu2021promoting,deng2021galerkin,callaham2021nonlinear,Callaham2022jfm,Callaham2022scienceadvances}, or other techniques such as reservoir computing~\cite{pathak2017using,pathak2018model}.  
When SINDy is used to learn the reduced-order model in Eq.~{eq:ROM}, this is referred to as Galerkin \emph{regression}~\cite{Loiseau2017jfm}, and this approach has been applied to a wide variety of problems in fluid dynamics~\cite{Loiseau2017jfm,Loiseau2018jfm,Deng2020JFM,Loiseau2020tcfd,deng2021galerkin,callaham2021nonlinear,Callaham2022jfm,Callaham2022scienceadvances}, electroconvection~\cite{guan2020sparse}, and plasmas~\cite{kaptanoglu2020physics,kaptanoglu2021promoting}.
It is also possible to combine SINDy for reduced-order modeling with autoencoders for dimensionality reduction, as in Champion et al.~\cite{Champion2019pnas}.  
Further, implicit kernel learning can be used to identify interpretable models when the state dimension is larger~\cite{baddoo2022LANDO}. 
Another important avenue of research, called \emph{lift and learn}~\cite{peherstorfer2016data,qian2020lift}, discovers lifting transformations so that more complex nonlinearities may be written as quadratic dynamics in a higher-dimensional coordinate system.  
Transformers have recently been used for operator learning in PDEs~\cite{cao2021choose,li2022transformer}.

\section{Numerical Methods and Learning Solution Operators}\label{Sec:Sec3}

The previous sections have focused on learning new PDEs from data or learning better coordinate systems in which to represent the PDEs.  A third major avenue of research, discussed here, involves improving existing numerical methods for solving PDEs with machine learning, including learning the solution operator directly.  

\subsection*{Operator inversion:  Green's functions and eigenfunction expansions}
For PDEs that are linear in $u(x,t)$ and its derivatives, i.e $N\rightarrow{\cal L}$, there are number of analytic techniques that have been historically developed in order to represent the solution.  For nonlinear operators $N(\cdot)$, one typically relies on numerical methods to approximate the solution.  Many modern deep learning methods are inspired by analytic techniques developed for linear operators for learning nonlinear operators.  To be more specific, consider the time-independent linear operator 
\begin{equation}
   {\cal L}u = f(x) .
   \label{eq:L}
\end{equation}
This problem has a rich and significant history in mathematical physics as quantum mechanics, electrodynamics and elasticity, for instance, all rely on understanding linear operators $ {\cal L}$~\cite{reed1980}. Indeed, Sturm-Liouville theory was a unifying and foundational theoretical advancement for understanding the underlying description of many physics-based problems of mathematical physics described by special functions (Bessel, Leguerre, Legendre, Hermite, etc).

The mathematical foundations for solving linear PDEs is to determine the inverse of ${\cal L}$.  Two methods, which are intimately connected, have traditionally been developed to represent the inverse operator ${\cal L}^{-1}$:  Green's functions and eigenfunction expansions.  For the Green's function, one considers the associated problem ${\cal L}^\dag G(x,\xi)= \delta(\xi)$ where ${\cal L}^\dag$ is the adjoint of ${\cal L}$, $\delta(\cdot)$ is the Dirac delta function and $\xi\in[0,L]$.  Taking the inner product of both sides of (\ref{eq:L}) with respect to $G(\xi,x)$ results in the solution
\begin{equation}
   u(x,t)= {\cal L}^{-1} f = \langle G | f \rangle = \int_D G(\xi,x) f(\xi) d\xi .
   \label{eq:green}
\end{equation}
Thus the inverse of the differential operator ${\cal L}$ is, not surprisingly, an integration over the Green's function which acts as kernel function.  Alternatively, eigenfunction expansions are based upon the associated spectral (eigenvalue) problem ${\cal L} \phi_n= \lambda_n \phi_n$ where $\phi_n$ and $\lambda_n$ are the eigenfunctions and eigenvalues respectively.  The success of the Sturm-Liouville theory is based upon a linear operator for which linear superposition holds, thus any solution can be represented by a sum of its eigenfunctions $u(x,t) =\sum b_n \phi_n$.  Inserting this solution form into (\ref{eq:L}) and taking inner products yields the solution form
\begin{equation}
   u(x,t)={\cal L}^{-1} f = \sum \frac{ \langle \phi_n | f \rangle}{\lambda_n} \phi_n.
   \label{eq:eigenfunction}
\end{equation}
Thus the solution is represented as a projection into the orthonormal coordinate space of the eigenfunctions.

The representations (\ref{eq:green}) and (\ref{eq:eigenfunction}) can be shown to be equivalent~\cite{courant2008methods}.  As noted, the difference in their representations of the solution have been used in deep learning to motivate learning for nonlinear operators.  Specifically, the Green's function representation has led to neural operators~\cite{li2020neural,li2020multipole,li2020fourier,kovachki2021neural}, DeepGreen~\cite{gin2020deepgreen} and DeepOnet~\cite{lu2019deeponet,lu2021learning} architectures, while the eigenfunction expansion solution is the basis of learning Koopman operators and their spectral representation~\cite{Mezic2004,Mezic2005nd,Brunton2021koopman}.  Like the Green's function and eigenfunction representations, they are alternative approaches for learning the inverse of an operator, in this case the inversion of the nonlinear operator $N^{-1}(\cdot)$.  The significant difficulty encountered for this inversion is that linear superposition no longer holds, thus undermining the creation of an underlying theoretical construct and guarantee for representation of the solution.  However, deep learning architectures can leverage training data to build accurate representations of the operator inversion.   The universal approximation capabilities of neural networks is well known which make them ideal for approximating the continuous functions associated with solutions of the PDE.

\subsection*{Operator learning and kernel methods} In the original paper advocating the construction of
{\em Neural Operators}, the Green's function representation of the solution motivates the proposed mappings between function spaces, thus allowing for the approximation of operators $N(\cdot)$ which encode governing equations and physics~\cite{li2020neural,li2020multipole,li2020fourier,kovachki2021neural}.  Thus neural operators leverage integral kernel representation in their approximation of the operator.  For instance, neural operators can make explicit use of multi-pole~\cite{li2020multipole} and Fourier~\cite{li2020fourier} kernels in order to construct operator representations.  Thus nonlocal representations of the solution are parametrized by the integral operator.  Recall that learning a nonlinear operator ${\cal G}(\cdot)$ will be equivalent to learning the inverse of the PDE evolution (\ref{eq:PDE}).  Thus kernel operators are intuitively appealing for the construction of the nonlinear operator based upon their connection to the linear kernel inversion (\ref{eq:green}).  The overall representation of the operator is a trained neural network
${\bf G}={\bf f}_{\boldsymbol{\theta}}$
where individual layers of the neural network are constructed from a learned integral representations that are updated according to the following kernel-based representation
\begin{equation}
{v}_{k+1} ({x}) = \sigma_{k+1} \left(  {W}_t {v}_k + \int_{{D}_k} K^{(k)} ({ x}, {y}) v_k(y) d\nu_k (y) + b_k(x)  \right)
\end{equation}
where $\nu_k$ is a Lebesgue measure on $\mathbb{R}^{d_t}$.  The kernel $K^{(k)} ({ x}, {y})$ is typically chosen to leverage advantageous representations, such as the multi-pole or Fourier kernels.  Thus each layer of the network is trained using a physics-inspired concept of an integral (inverse) representation of the PDE dynamics.  Thus instead of constructing a Green's function kernel, which can technically only be done with a linear operator ${\cal L}$, the kernel representation is used to train a representation of the inverse operator $N^{-1}(\cdot)$.
Rigorous estimates of the convergence rates and computational costs for learning such linear operators can now be derived rigorously~\cite{de2021convergence,de2022cost,mollenhauer2022learning}.

\begin{figure}[t]
 \centering
\begin{overpic}[width=0.8\textwidth]{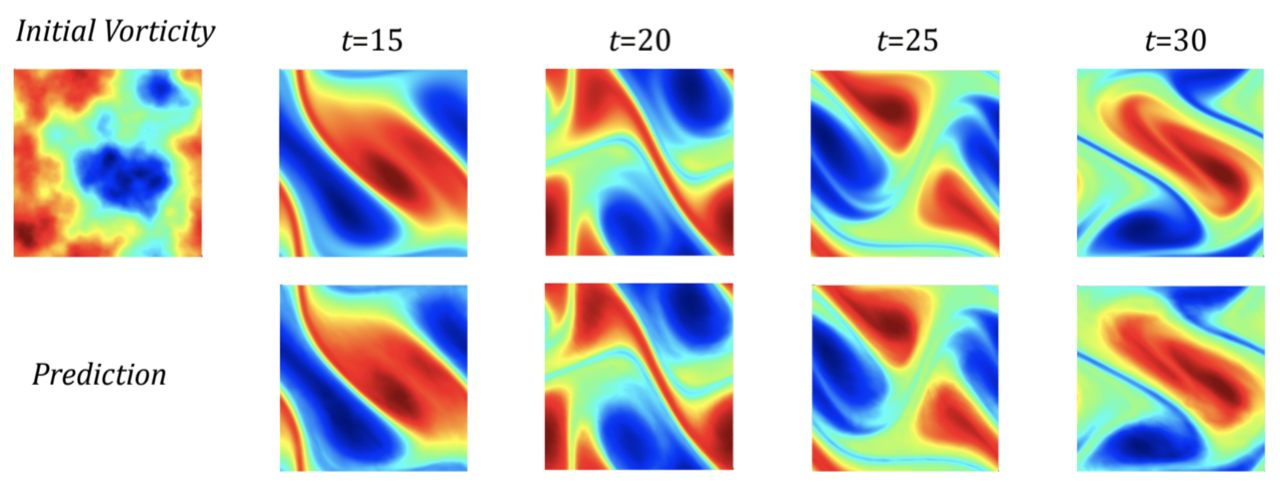}
\end{overpic}
\vspace*{-0.1in}
\caption{Zero-shot super-resolution: Vorticity field of the solution to the two-dimensional Navier-Stokes equation with viscosity $10^4$ ($Re=O(200)$); Ground truth on top and prediction on bottom. The model is trained on data that is discretized on a uniform 64 × 64 spatial grid and on a 20-point uniform temporal grid. The model is evaluated with a different initial condition that is discretized on a uniform 256 × 256 spatial grid and a 80-point uniform temporal grid ({\em From Kovachki et al~\cite{kovachki2021neural}} ).}
\label{fig:neuralO}
\end{figure}

More broadly, neural operators  generalize standard feed-forward neural networks to learn mappings between infinite-dimensional spaces of functions defined on bounded domains of $\mathbb{R}^d$. The non-local component of the architecture is instantiated through either a parameterized integral operator or through multiplication in the spectral domain (which is a specific form of the kernel in the integral operator).  Once trained, neural operators have the property of being discretization invariant: sharing the same network parameters between different discretizations of the underlying functional data. Thus it is a mesh free method, as shown in Fig.~\ref{fig:neuralO} on the Navier-Stokes equation.

On a more foundational level, Chen and Chen~\cite{chen1995universal} developed a proof that neural networks with a single hidden layer can approximate accurately any nonlinear continuous operator.  Thus a nonlinear operator is learned mapping from functions to functions.  In practice, this is a highly impactful theory as it provides guarantees on the construction of an operator which contains information about the physics and dynamics of the system.  The theorem of Chen and Chen is the basis of the DeepOnet method of Lu et al~\cite{lu2019deeponet,lu2021learning} ({\em DeepOnet}) as well as the neural operators of Kovachki et al~\cite{kovachki2021neural}.  The original work of Chen and Chen~\cite{chen1995universal} construct a universal approximation proof.  The theorem provides a theoretical bounds on the ability of a neural network to approximate the operator ${\cal G}(\cdot)$.  It also highlights the construction of two neural networks so that it can be more compactly represented as
\begin{equation}
  \left|
   {\cal G}({\bf u})({\bf y}) -  {\bf f}_{\boldsymbol{\theta}_1} ({\bf u}) \cdot  \tilde{\bf f}_{\boldsymbol{\theta}_2} ({\bf y})
  \right|  < \epsilon
\end{equation}
when considering the discretized representation of $u(x)\rightarrow {\bf u}$ and new measurement (function evaluation) locations $y\rightarrow {\bf y}$.  
The two simultaneously trained networks are the branch network ${\bf f}_{\boldsymbol{\theta}_1} ({\bf u})$ and the trunk network $\tilde{\bf f}_{\boldsymbol{\theta}_2} ({\bf y})$.  

\begin{figure}[t]
 \centering
\begin{overpic}[width=0.85\textwidth]{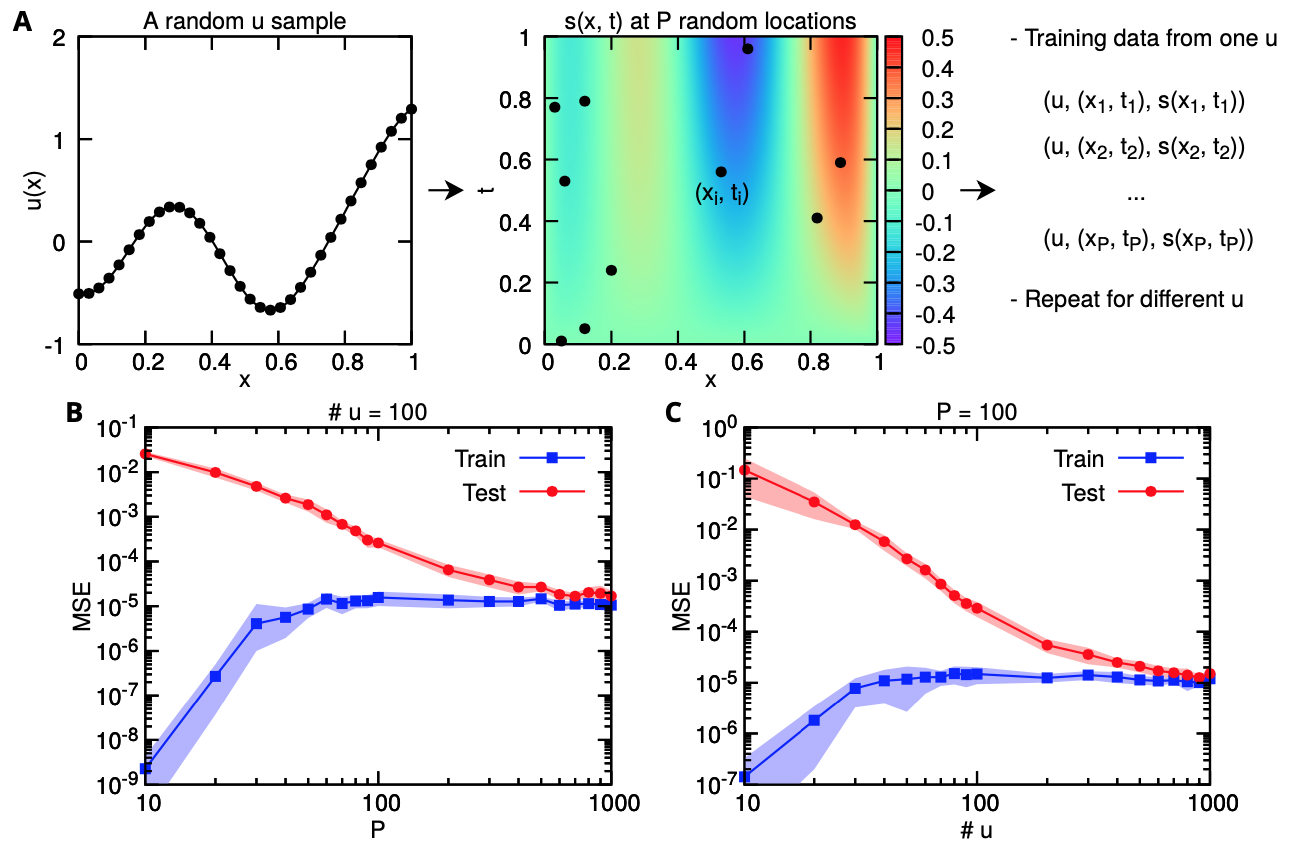}
\end{overpic}
\vspace*{-0.1in}
\caption{Learning a reaction-diffusion with DeepOnet. (A) (left) An example of a random sample of the input function $u(x)$. (middle) The corresponding output function $s(x, t)$ at P different $(x, t)$ locations. (right) Pairing of inputs and outputs at the training data points. The total number of training data points is the product of $P$ times the number of samples of $u$. (B) Training error (blue) and test error (red) for different values of the number of random points $P$ when 100 random $u$ samples are used. (C) Training error (blue) and test error (red) for different number of u samples when $P = 100$. The shaded regions denote one-standard-derivation ({\em From Lu et al~\cite{lu2019deeponet}} ).}
\label{fig:DeepO2}
\end{figure}

Mathematically, the concept is quite simple.  Given a number of measurement (sensor) locations $x_k$ (usually selected from a computational grid) which prescribes the input function $u_k=u(x_k)$, a vector of training input data can be constructed ${\bf u}$.  The input data has a corresponding output data ${\cal G}({\bf u})$.
In addition, training data mapping selections of random measurement points ${\bf y}$ to the output ${\cal G}({\bf u}({\bf y})$ is required.   Thus the input functions ${\bf u}$ are encoded in a separate network than the location variables ${\bf y}$.  These are merged at the end as shown in the universal approximation proof of Chen and Chen~\cite{chen1995universal}.
Figure~\ref{fig:DeepO2} shows the results of training from the original DeepOnet paper of  Lu et al~\cite{lu2019deeponet,lu2021learning,li2020fourier} on reaction-diffusion system.   DeepOnets also can achieve small generalization errors by employing inductive biases.  Remarkably, exponential convergence is observed in the deep learning algorithm.  
 
So although both neural operators and DeepOnets accomplish the same goal, they do so with significantly different architectures.   Neural operators exploit the kernel structure of generic operators while DeepOnets train by separating the input function from the spatial locations.  Both have achieved promising results, highlighting the fact that the learning of operators can potentially allow for mesh-free models of physics systems.  Of course, in order for this to actually be viable in practice, exceptional training data that resolves all scales should be employed in training.   Figure~\ref{fig:neuralO} highlights the results from Kovachki et al~\cite{kovachki2021neural} where neural operators are used to model fluid flows.  

\subsection*{Accelerating numerical solutions}
Finally, there are several emerging approaches to directly accelerate the numerical computation of the solutions to PDEs.  
For example, the Navier-Stokes equations for fluid flows are notoriously challenging to simulate at all resolutions because of the large degree of scale-separation in space and time.  
Improving the comptutational scaling, accuracy, and efficiency of numerical schemes is an important topic of modern machine-learning-enabled scientific computing~\cite{vinuesa2022enhancing}.  
For example, Bar-Sinai {\it et al.}~\cite{bar2019learning} developed a deep learning approach to improve the estimation of spatial derivatives on coarse grids, outperforming traditional finite-difference methods. 
Stevens and Colonius~\cite{stevens2020enhancement} developed a related approach that improved upon fifth-order finite-difference schemes for shock-capturing simulations.  
These approaches solve a similar task as the neural operator approaches discussed earlier~\cite{li2020fourier,li2020multipole,li2020neural}, which seek to improve simulations on coarser meshes.  
More classically, data-driven methods based on the ROMs developed earlier make it possible to learn more effective collocation points, resulting in discrete empirical interpolation methods for PDEs~\cite{Barrault2004crm,Chaturantabut2010siamjsc}. 
Other approaches, such as FiniteNet~\cite{stevens2020finitenet} leverage long-short term memory (LSTM) networks to improve the simulation efficiency of PDEs. 
Machine learning is also improving the conditioning of flow solvers and the computation of inflow boundary conditions, for example with transformers~\cite{yousif2022transformer}. 

\begin{figure}
    \centering    \includegraphics{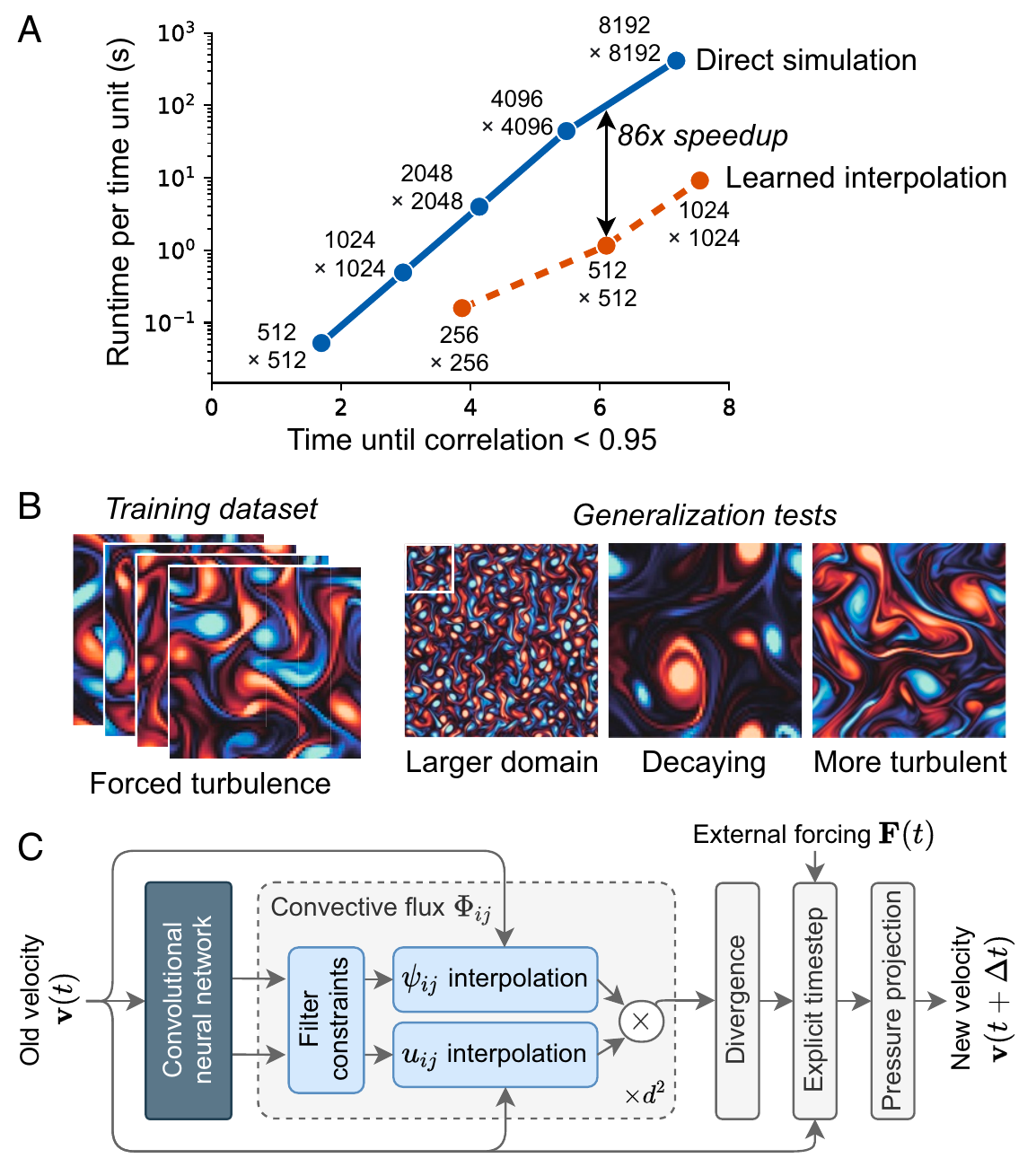}
    \caption{Machine learned interpolation from coarse-grained to high-resolution flow fields, reproduced from Kochkov et al.~\cite{kochkov2021machine}.}
    \label{fig:Kochkov}
\end{figure}

Recently, Kochkov~{\it et al.}~\cite{kochkov2021machine} developed a deep learning correction for the two-dimensional Kolmogorov flow, showing that it is possible to simulate on a much coarser grid than is traditionally possible (e.g., approximately 10 times coarser in each dimension).  Figure~\ref{fig:Kochkov} shows the performance and architecture. 
This approach is morally similar to super-resolution efforts, which have gained considerable attention in fluid dynamics applications~\cite{fukami2019super,fukami2021machine,fukami2021robust}. 
Extending these various approaches to three dimensions and more complex flows is necessary for these methods to gain wide adoption, and this represents an important and active area of current research. 

\section{Discussion}\label{Sec:Discussion}
In this perspective, we have explored how emerging techniques in machine learning are enabling major advances in the field of partial differential equations.  In particular, we have summarized efforts to 1) discover new PDEs and coarse-grained closure models from data, 2) to uncover new coordinate systems in which the PDE and its solution become simpler, and 3) to directly learn solution operators and other techniques to accelerate numerics.  In every case, despite significant progress, there are several ongoing challenges and opportunities for development.  

In the field of discovery and coarse-graining, there are several avenues of ongoing research.  Preliminary results show that it is possible to learn new physical mechanisms and closure models, mainly in fluid systems.  There is a tremendous opportunity to refine and leverage these new closure models to accelerate simulations of turbulent fluid systems to enable their use in a diverse range of applications and technologies.   
Moreover, there are many new fields where this approach might be applied: neuroscience, epidemiology, active matter, non-Newtonian fluids, among others.  
In addition, there is an opportunity to incorporate partial knowledge of the physics, including symmetries and invariances.  The dual of this, is that given a new discovered PDE, it may be possible to relate this to a new conservation or invariance.  
In any of these situations, when a PDE is uncovered, it is possible to automatically cluster the dynamics in space and time by what terms in the PDE are in a dominant balance with eachother.  Similarly, it may be possible to identify the controlling nondimensional parameters that determine the bifurcation structure of the system.  

Even when a PDE is known, from first principles or from data-driven learning algorithms, the presence of nonlinearity makes it so that there are no generic solution techniques.  
We have seen that advances in Koopman operator theory are making it possible to learn new coordinate systems in which nonlinear systems become linear.  For example, the Cole-Hopf transformation may be seen as a Koopman coordinate transformation in which case the nonlinear Burgers' equation maps into the linear heat equation.  
There are many opportunities to discovery similar coordinate transformations for more complex systems, such as the Navier-Stokes equations.  
In addition to learning linearizing transformations, it may be possible to relax this stringent constraint, and instead learn transformations into a coordinate system where the dynamics are simplified, with asymptotic or perturbative nonlinearities.  This is related to normal form theory, where it may be possible to dramatically simplify the dynamics with a much less complex coordinate transfomrmation.  

Finally, there are several efforts underway to accelerate numerics associated with solving PDEs, as well as to approximate the solution operators directly.  
The universal approximation capabilities of neural networks make them particularly useful for representing the solutions to PDEs, which may be arbitrarily complex.  
Understanding how these solution operators vary with system parameters is an important avenue of ongoing research~\cite{pan2023neural}.  
Similarly, machine learning may be used to accelerate traditional scientific computing workflows, for example by flexible super-resolution or learning of improved solution stencils.  
However, here are several challenges with these approaches, foremost the fact that traditional numerical algorithms are extremely mature and scaleable, so that machine learning solutions are expected to compete with decades of progress. 

In all of the cases explored in this perspective, progress will be accelerated by a diverse and robust set of benchmark problems with which to assess new solutions~\cite{takamoto2022pdebench}. In addition, we must stress that these techniques are primarily tools to be used by human experts for scientific discovery.  
In the past, many advances have been driven in the field of fluid mechanics~\cite{Brunton2020arfm}, and this is likely to continue.  For example, understanding sensitivities with resolvent analysis~\cite{sharma2013coherent}, using physics informed neural networks (PINNs)~\cite{Raissi2019jcp} for RANS modeling~\cite{eivazi2022physics}, and using wall measurements to estimate turbulent flow fields~\cite{guemes2021coarse} are all exciting avenues of research.  
Interestingly, there are also efforts to understand neural networks using techniques from PDEs~\cite{burger2021connections}.

Although there is a desire for automated machine learning algorithms, when applied to science and engineering applications, this is still primarily a human endeavor. However, progress in the field of PDEs, enabled by machine learning, is undeniable.  
Despite this progress, there is still much we don’t know about PDEs.  For example, it is unknown whether or not all solutions of the incompressible fluid flow equations even remain bounded in finite time, making it one of the ``Millennium Prize” problems.  Our limitations in our understanding of PDEs is nicely summarized by Richard Feynman~\cite{Feynman2013book}:
\begin{quote}
    “The next great era of awakening of human intellect may well produce a method of understanding the qualitative content of equations. Today we cannot. Today we cannot see that the water flow equations contain such things as the barber pole structure of turbulence that one sees between rotating cylinders. Today we cannot see whether Schrodinger's equation contains frogs, musical composers, or morality--or whether it does not. We cannot say whether something beyond it like God is needed, or not. And so we can all hold strong opinions either way.”
\end{quote}

\section*{Acknowledgements}
The authors acknowledge support from the National Science Foundation AI Institute in Dynamic Systems
(grant number 2112085).  SLB acknowledges support from the Army Research Office (ARO W911NF-19-1-0045). 

\begin{spacing}{.88}
\setlength{\bibsep}{2.pt}

\begin{thebibliography}{140}
\providecommand{\natexlab}[1]{#1}
\providecommand{\url}[1]{\texttt{#1}}
\expandafter\ifx\csname urlstyle\endcsname\relax
  \providecommand{\doi}[1]{doi: #1}\else
  \providecommand{\doi}{doi: \begingroup \urlstyle{rm}\Url}\fi

\bibitem[Ablowitz and Segur(1981)]{ablowitz1981solitons}
M.~J. Ablowitz and H.~Segur.
\newblock \emph{Solitons and the inverse scattering transform}.
\newblock SIAM, 1981.

\bibitem[Ablowitz et~al.(1974)Ablowitz, Kaup, Newell, and
  Segur]{ablowitz1974inverse}
M.~J. Ablowitz, D.~J. Kaup, A.~C. Newell, and H.~Segur.
\newblock The inverse scattering transform-fourier analysis for nonlinear
  problems.
\newblock \emph{Studies in applied mathematics}, 53\penalty0 (4):\penalty0
  249--315, 1974.

\bibitem[Abraham et~al.(1988)Abraham, Marsden, and Ratiu]{MarsdenMTAA}
R.~Abraham, J.~E. Marsden, and T.~Ratiu.
\newblock \emph{Manifolds, Tensor Analysis, and Applications}, volume~75 of
  \emph{Applied Mathematical Sciences}.
\newblock Springer-Verlag, 1988.

\bibitem[Ahmed et~al.(2021)Ahmed, Pawar, San, Rasheed, Iliescu, and
  Noack]{ahmed2021closures}
S.~E. Ahmed, S.~Pawar, O.~San, A.~Rasheed, T.~Iliescu, and B.~R. Noack.
\newblock On closures for reduced order models---a spectrum of first-principle
  to machine-learned avenues.
\newblock \emph{Physics of Fluids}, 33\penalty0 (9):\penalty0 091301, 2021.

\bibitem[Alves and Fiuza(2020)]{alves2020data}
E.~P. Alves and F.~Fiuza.
\newblock Data-driven discovery of reduced plasma physics models from
  fully-kinetic simulations.
\newblock \emph{arXiv preprint arXiv:2011.01927}, 2020.

\bibitem[Atkinson(2020)]{atkinson2020bayesian}
S.~Atkinson.
\newblock Bayesian hidden physics models: Uncertainty quantification for
  discovery of nonlinear partial differential operators from data.
\newblock \emph{arXiv preprint arXiv:2006.04228}, 2020.

\bibitem[Baddoo et~al.(2022)Baddoo, Herrmann, McKeon, and
  Brunton]{baddoo2022LANDO}
P.~J. Baddoo, B.~Herrmann, B.~J. McKeon, and S.~L. Brunton.
\newblock Kernel learning for robust dynamic mode decomposition: Linear and
  nonlinear disambiguation optimization (lando).
\newblock \emph{Proceedings of the Royal Society A}, 478\penalty0
  (2260):\penalty0 20210830, 2022.

\bibitem[Bakarji and Tartakovsky(2020)]{bakarji2020data}
J.~Bakarji and D.~M. Tartakovsky.
\newblock Data-driven discovery of coarse-grained equations.
\newblock \emph{arXiv preprint arXiv:2002.00790}, 2020.

\bibitem[Bar-Sinai et~al.(2019)Bar-Sinai, Hoyer, Hickey, and
  Brenner]{bar2019learning}
Y.~Bar-Sinai, S.~Hoyer, J.~Hickey, and M.~P. Brenner.
\newblock Learning data-driven discretizations for partial differential
  equations.
\newblock \emph{Proceedings of the National Academy of Sciences}, 116\penalty0
  (31):\penalty0 15344--15349, 2019.

\bibitem[Barrault et~al.(2004)Barrault, Maday, Nguyen, and
  Patera]{Barrault2004crm}
M.~Barrault, Y.~Maday, N.~C. Nguyen, and A.~T. Patera.
\newblock An `empirical interpolation'method: application to efficient
  reduced-basis discretization of partial differential equations.
\newblock \emph{Comptes Rendus Mathematique}, 339\penalty0 (9):\penalty0
  667--672, 2004.

\bibitem[Beetham and Capecelatro(2020)]{beetham2020formulating}
S.~Beetham and J.~Capecelatro.
\newblock Formulating turbulence closures using sparse regression with embedded
  form invariance.
\newblock \emph{Physical Review Fluids}, 5\penalty0 (8):\penalty0 084611, 2020.

\bibitem[Beetham et~al.(2021)Beetham, Fox, and Capecelatro]{beetham2021sparse}
S.~Beetham, R.~O. Fox, and J.~Capecelatro.
\newblock Sparse identification of multiphase turbulence closures for coupled
  fluid--particle flows.
\newblock \emph{Journal of Fluid Mechanics}, 914, 2021.

\bibitem[Benner et~al.(2015)Benner, Gugercin, and
  Willcox]{Benner2015siamreview}
P.~Benner, S.~Gugercin, and K.~Willcox.
\newblock A survey of projection-based model reduction methods for parametric
  dynamical systems.
\newblock \emph{SIAM review}, 57\penalty0 (4):\penalty0 483--531, 2015.

\bibitem[Benner et~al.(2020)Benner, Goyal, Kramer, Peherstorfer, and
  Willcox]{benner2020operator}
P.~Benner, P.~Goyal, B.~Kramer, B.~Peherstorfer, and K.~Willcox.
\newblock Operator inference for non-intrusive model reduction of systems with
  non-polynomial nonlinear terms.
\newblock \emph{Computer Methods in Applied Mechanics and Engineering},
  372:\penalty0 113433, 2020.

\bibitem[Berkooz et~al.(1993)Berkooz, Holmes, and Lumley]{Berkooz1993arfm}
G.~Berkooz, P.~Holmes, and J.~Lumley.
\newblock The proper orthogonal decomposition in the analysis of turbulent
  flows.
\newblock \emph{Ann.\ Rev.\ Fluid Mech.}, 25:\penalty0 539--575, 1993.

\bibitem[Bongard and Lipson(2007)]{Bongard2007pnas}
J.~Bongard and H.~Lipson.
\newblock Automated reverse engineering of nonlinear dynamical systems.
\newblock \emph{Proceedings of the National Academy of Sciences}, 104\penalty0
  (24):\penalty0 9943--9948, 2007.

\bibitem[Brandstetter et~al.(2022{\natexlab{a}})Brandstetter, Berg, Welling,
  and Gupta]{brandstetter2022clifford}
J.~Brandstetter, R.~v.~d. Berg, M.~Welling, and J.~K. Gupta.
\newblock Clifford neural layers for pde modeling.
\newblock \emph{arXiv preprint arXiv:2209.04934}, 2022{\natexlab{a}}.

\bibitem[Brandstetter et~al.(2022{\natexlab{b}})Brandstetter, Welling, and
  Worrall]{brandstetter2022lie}
J.~Brandstetter, M.~Welling, and D.~E. Worrall.
\newblock Lie point symmetry data augmentation for neural pde solvers.
\newblock In \emph{International Conference on Machine Learning}, pages
  2241--2256. PMLR, 2022{\natexlab{b}}.

\bibitem[Brandstetter et~al.(2022{\natexlab{c}})Brandstetter, Worrall, and
  Welling]{brandstetter2022message}
J.~Brandstetter, D.~Worrall, and M.~Welling.
\newblock Message passing neural pde solvers.
\newblock \emph{arXiv preprint arXiv:2202.03376}, 2022{\natexlab{c}}.

\bibitem[Brezis and Browder(1998)]{brezis1998partial}
H.~Brezis and F.~Browder.
\newblock Partial differential equations in the 20th century.
\newblock \emph{Advances in mathematics}, 135\penalty0 (1):\penalty0 76--144,
  1998.

\bibitem[Brunton and Kutz(2022)]{Brunton2022book}
S.~L. Brunton and J.~N. Kutz.
\newblock \emph{Data-Driven Science and Engineering: Machine Learning,
  Dynamical Systems, and Control}.
\newblock Cambridge University Press, 2nd edition, 2022.

\bibitem[Brunton et~al.(2016)Brunton, Proctor, and Kutz]{Brunton2016pnas}
S.~L. Brunton, J.~L. Proctor, and J.~N. Kutz.
\newblock Discovering governing equations from data by sparse identification of
  nonlinear dynamical systems.
\newblock \emph{Proceedings of the National Academy of Sciences}, 113\penalty0
  (15):\penalty0 3932--3937, 2016.

\bibitem[Brunton et~al.(2020)Brunton, Noack, and Koumoutsakos]{Brunton2020arfm}
S.~L. Brunton, B.~R. Noack, and P.~Koumoutsakos.
\newblock Machine learning for fluid mechanics.
\newblock \emph{Annual Review of Fluid Mechanics}, 52:\penalty0 477--508, 2020.

\bibitem[Brunton et~al.(2021)Brunton, Budi{\v{s}}i{\'c}, Kaiser, and
  Kutz]{Brunton2021koopman}
S.~L. Brunton, M.~Budi{\v{s}}i{\'c}, E.~Kaiser, and J.~N. Kutz.
\newblock Modern {K}oopman theory for dynamical systems.
\newblock \emph{arXiv preprint arXiv:2102.12086}, 2021.

\bibitem[Brunton et~al.(2022)Brunton, Budi{\v{s}}i{\'c}, Kaiser, and
  Kutz]{Brunton2022siamreview}
S.~L. Brunton, M.~Budi{\v{s}}i{\'c}, E.~Kaiser, and J.~N. Kutz.
\newblock Modern {K}oopman theory for dynamical systems.
\newblock \emph{SIAM Review}, 64\penalty0 (2):\penalty0 229--340, 2022.

\bibitem[Budi{\v{s}}i{\'c} et~al.(2012)Budi{\v{s}}i{\'c}, Mohr, and
  Mezi{\'c}]{Budivsic2012chaos}
M.~Budi{\v{s}}i{\'c}, R.~Mohr, and I.~Mezi{\'c}.
\newblock Applied {K}oopmanism a).
\newblock \emph{Chaos: An Interdisciplinary Journal of Nonlinear Science},
  22\penalty0 (4):\penalty0 047510, 2012.

\bibitem[Burger et~al.(2021)Burger, RUTHOTTO, OSHER,
  et~al.]{burger2021connections}
M.~Burger, L.~RUTHOTTO, S.~OSHER, et~al.
\newblock Connections between deep learning and partial differential equations.
\newblock \emph{European Journal of Applied Mathematics}, 32\penalty0
  (3):\penalty0 395--396, 2021.

\bibitem[Callaham et~al.(2021{\natexlab{a}})Callaham, Koch, Brunton, Kutz, and
  Brunton]{callaham2021learning}
J.~L. Callaham, J.~V. Koch, B.~W. Brunton, J.~N. Kutz, and S.~L. Brunton.
\newblock Learning dominant physical processes with data-driven balance models.
\newblock \emph{Nature communications}, 12\penalty0 (1):\penalty0 1--10,
  2021{\natexlab{a}}.

\bibitem[Callaham et~al.(2021{\natexlab{b}})Callaham, Loiseau, Rigas, and
  Brunton]{callaham2021nonlinear}
J.~L. Callaham, J.-C. Loiseau, G.~Rigas, and S.~L. Brunton.
\newblock Nonlinear stochastic modelling with langevin regression.
\newblock \emph{Proceedings of the Royal Society A}, 477\penalty0
  (2250):\penalty0 20210092, 2021{\natexlab{b}}.

\bibitem[Callaham et~al.(2022{\natexlab{a}})Callaham, Brunton, and
  Loiseau]{Callaham2022jfm}
J.~L. Callaham, S.~L. Brunton, and J.-C. Loiseau.
\newblock On the role of nonlinear correlations in reduced-order modeling.
\newblock \emph{Journal of Fluid Mechanics}, 938\penalty0 (A1),
  2022{\natexlab{a}}.

\bibitem[Callaham et~al.(2022{\natexlab{b}})Callaham, Rigas, Loiseau, and
  Brunton]{Callaham2022scienceadvances}
J.~L. Callaham, G.~Rigas, J.-C. Loiseau, and S.~L. Brunton.
\newblock An empirical mean-field model of symmetry-breaking in a turbulent
  wake.
\newblock \emph{Science Advances}, 8\penalty0 (eabm4786), 2022{\natexlab{b}}.

\bibitem[Cao(2021)]{cao2021choose}
S.~Cao.
\newblock Choose a transformer: Fourier or galerkin.
\newblock \emph{Advances in neural information processing systems},
  34:\penalty0 24924--24940, 2021.

\bibitem[Champion et~al.(2019)Champion, Lusch, Kutz, and
  Brunton]{Champion2019pnas}
K.~Champion, B.~Lusch, J.~N. Kutz, and S.~L. Brunton.
\newblock Data-driven discovery of coordinates and governing equations.
\newblock \emph{Proceedings of the National Academy of Sciences}, 116\penalty0
  (45):\penalty0 22445--22451, 2019.

\bibitem[Chaturantabut and Sorensen(2010)]{Chaturantabut2010siamjsc}
S.~Chaturantabut and D.~C. Sorensen.
\newblock Nonlinear model reduction via discrete empirical interpolation.
\newblock \emph{SIAM Journal on Scientific Computing}, 32\penalty0
  (5):\penalty0 2737--2764, 2010.

\bibitem[Chen and Chen(1995)]{chen1995universal}
T.~Chen and H.~Chen.
\newblock Universal approximation to nonlinear operators by neural networks
  with arbitrary activation functions and its application to dynamical systems.
\newblock \emph{IEEE Transactions on Neural Networks}, 6\penalty0 (4):\penalty0
  911--917, 1995.

\bibitem[Cole(1951)]{cole51}
J.~D. Cole.
\newblock On a quasi-linear parabolic equation occurring in aerodynamics.
\newblock \emph{Quart. Appl. Math.}, 9:\penalty0 225--236, 1951.

\bibitem[Courant and Hilbert(2008)]{courant2008methods}
R.~Courant and D.~Hilbert.
\newblock \emph{Methods of mathematical physics: partial differential
  equations}.
\newblock John Wiley \& Sons, 2008.

\bibitem[Cover(1965)]{cover1965geometrical}
T.~M. Cover.
\newblock Geometrical and statistical properties of systems of linear
  inequalities with applications in pattern recognition.
\newblock \emph{IEEE transactions on electronic computers}, \penalty0
  (3):\penalty0 326--334, 1965.

\bibitem[Cranmer et~al.(2020{\natexlab{a}})Cranmer, Greydanus, Hoyer,
  Battaglia, Spergel, and Ho]{cranmer2020lagrangian}
M.~Cranmer, S.~Greydanus, S.~Hoyer, P.~Battaglia, D.~Spergel, and S.~Ho.
\newblock Lagrangian neural networks.
\newblock \emph{arXiv preprint arXiv:2003.04630}, 2020{\natexlab{a}}.

\bibitem[Cranmer et~al.(2020{\natexlab{b}})Cranmer, Sanchez-Gonzalez,
  Battaglia, Xu, Cranmer, Spergel, and Ho]{cranmer2020discovering}
M.~Cranmer, A.~Sanchez-Gonzalez, P.~Battaglia, R.~Xu, K.~Cranmer, D.~Spergel,
  and S.~Ho.
\newblock Discovering symbolic models from deep learning with inductive biases.
\newblock \emph{arXiv preprint arXiv:2006.11287}, 2020{\natexlab{b}}.

\bibitem[Cranmer et~al.(2019)Cranmer, Xu, Battaglia, and
  Ho]{cranmer2019learning}
M.~D. Cranmer, R.~Xu, P.~Battaglia, and S.~Ho.
\newblock Learning symbolic physics with graph networks.
\newblock \emph{arXiv preprint arXiv:1909.05862}, 2019.

\bibitem[De~Haan et~al.(2020)De~Haan, Weiler, Cohen, and Welling]{de2020gauge}
P.~De~Haan, M.~Weiler, T.~Cohen, and M.~Welling.
\newblock Gauge equivariant mesh cnns: Anisotropic convolutions on geometric
  graphs.
\newblock \emph{arXiv preprint arXiv:2003.05425}, 2020.

\bibitem[De~Hoop et~al.(2022)De~Hoop, Huang, Qian, and Stuart]{de2022cost}
M.~De~Hoop, D.~Z. Huang, E.~Qian, and A.~M. Stuart.
\newblock The cost-accuracy trade-off in operator learning with neural
  networks.
\newblock \emph{arXiv preprint arXiv:2203.13181}, 2022.

\bibitem[de~Hoop et~al.(2021)de~Hoop, Kovachki, Nelsen, and
  Stuart]{de2021convergence}
M.~V. de~Hoop, N.~B. Kovachki, N.~H. Nelsen, and A.~M. Stuart.
\newblock Convergence rates for learning linear operators from noisy data.
\newblock \emph{arXiv preprint arXiv:2108.12515}, 2021.

\bibitem[Deng et~al.(2020)Deng, Noack, Morzynski, and Pastur]{Deng2020JFM}
N.~Deng, B.~R. Noack, M.~Morzynski, and L.~R. Pastur.
\newblock Low-order model for successive bifurcations of the fluidic pinball.
\newblock \emph{Journal of fluid mechanics}, 884\penalty0 (A37), 2020.

\bibitem[Deng et~al.(2021)Deng, Noack, Morzy{\'n}ski, and
  Pastur]{deng2021galerkin}
N.~Deng, B.~R. Noack, M.~Morzy{\'n}ski, and L.~R. Pastur.
\newblock Galerkin force model for transient and post-transient dynamics of the
  fluidic pinball.
\newblock \emph{Journal of Fluid Mechanics}, 918, 2021.

\bibitem[Dissanayake and Phan-Thien(1994)]{dissanayake1994neural}
M.~Dissanayake and N.~Phan-Thien.
\newblock Neural-network-based approximations for solving partial differential
  equations.
\newblock \emph{communications in Numerical Methods in Engineering},
  10\penalty0 (3):\penalty0 195--201, 1994.

\bibitem[Duraisamy et~al.(2019)Duraisamy, Iaccarino, and
  Xiao]{Duraisamy2019arfm}
K.~Duraisamy, G.~Iaccarino, and H.~Xiao.
\newblock Turbulence modeling in the age of data.
\newblock \emph{Annual Reviews of Fluid Mechanics}, 51:\penalty0 357--377,
  2019.

\bibitem[Eivazi et~al.(2021)Eivazi, Guastoni, Schlatter, Azizpour, and
  Vinuesa]{eivazi2021recurrent}
H.~Eivazi, L.~Guastoni, P.~Schlatter, H.~Azizpour, and R.~Vinuesa.
\newblock Recurrent neural networks and koopman-based frameworks for temporal
  predictions in a low-order model of turbulence.
\newblock \emph{International Journal of Heat and Fluid Flow}, 90:\penalty0
  108816, 2021.

\bibitem[Eivazi et~al.(2022)Eivazi, Tahani, Schlatter, and
  Vinuesa]{eivazi2022physics}
H.~Eivazi, M.~Tahani, P.~Schlatter, and R.~Vinuesa.
\newblock Physics-informed neural networks for solving reynolds-averaged
  navier--stokes equations.
\newblock \emph{Physics of Fluids}, 34\penalty0 (7):\penalty0 075117, 2022.

\bibitem[Fasel et~al.(2022)Fasel, Kutz, Brunton, and
  Brunton]{fasel2022ensemble}
U.~Fasel, J.~N. Kutz, B.~W. Brunton, and S.~L. Brunton.
\newblock Ensemble-sindy: Robust sparse model discovery in the low-data,
  high-noise limit, with active learning and control.
\newblock \emph{Proceedings of the Royal Society A}, 478\penalty0
  (2260):\penalty0 20210904, 2022.

\bibitem[Feynman et~al.(2013)Feynman, Leighton, and Sands]{Feynman2013book}
R.~P. Feynman, R.~B. Leighton, and M.~Sands.
\newblock \emph{The Feynman Lectures on Physics}, volume~2.
\newblock Basic Books, 2013.

\bibitem[Fukami and Taira(2021)]{fukami2021robust}
K.~Fukami and K.~Taira.
\newblock Robust machine learning of turbulence through generalized buckingham
  pi-inspired pre-processing of training data.
\newblock In \emph{APS Division of Fluid Dynamics Meeting Abstracts}, pages
  A31--004, 2021.

\bibitem[Fukami et~al.(2019)Fukami, Fukagata, and Taira]{fukami2019super}
K.~Fukami, K.~Fukagata, and K.~Taira.
\newblock Super-resolution reconstruction of turbulent flows with machine
  learning.
\newblock \emph{Journal of Fluid Mechanics}, 870:\penalty0 106--120, 2019.

\bibitem[Fukami et~al.(2021)Fukami, Fukagata, and Taira]{fukami2021machine}
K.~Fukami, K.~Fukagata, and K.~Taira.
\newblock Machine-learning-based spatio-temporal super resolution
  reconstruction of turbulent flows.
\newblock \emph{Journal of Fluid Mechanics}, 909, 2021.

\bibitem[Gin et~al.(2021)Gin, Lusch, Brunton, and Kutz]{gin2021deep}
C.~Gin, B.~Lusch, S.~L. Brunton, and J.~N. Kutz.
\newblock Deep learning models for global coordinate transformations that
  linearise pdes.
\newblock \emph{European Journal of Applied Mathematics}, 32\penalty0
  (3):\penalty0 515--539, 2021.

\bibitem[Gin et~al.(2020)Gin, Shea, Brunton, and Kutz]{gin2020deepgreen}
C.~R. Gin, D.~E. Shea, S.~L. Brunton, and J.~N. Kutz.
\newblock Deepgreen: Deep learning of green's functions for nonlinear boundary
  value problems.
\newblock \emph{arXiv preprint arXiv:2101.07206}, 2020.

\bibitem[Guan et~al.(2021)Guan, Brunton, and Novosselov]{guan2020sparse}
Y.~Guan, S.~L. Brunton, and I.~Novosselov.
\newblock Sparse nonlinear models of chaotic electroconvection.
\newblock \emph{Royal Society Open Science}, 8\penalty0 (8):\penalty0 202367,
  2021.

\bibitem[G{\"u}emes et~al.(2021)G{\"u}emes, Discetti, Ianiro, Sirmacek,
  Azizpour, and Vinuesa]{guemes2021coarse}
A.~G{\"u}emes, S.~Discetti, A.~Ianiro, B.~Sirmacek, H.~Azizpour, and
  R.~Vinuesa.
\newblock From coarse wall measurements to turbulent velocity fields through
  deep learning.
\newblock \emph{Physics of Fluids}, 33\penalty0 (7):\penalty0 075121, 2021.

\bibitem[Gurevich et~al.(2019)Gurevich, Reinbold, and
  Grigoriev]{gurevich2019robust}
D.~R. Gurevich, P.~A. Reinbold, and R.~O. Grigoriev.
\newblock Robust and optimal sparse regression for nonlinear pde models.
\newblock \emph{Chaos: An Interdisciplinary Journal of Nonlinear Science},
  29\penalty0 (10):\penalty0 103113, 2019.

\bibitem[Holmes and Guckenheimer(1983)]{guckenheimer_holmes}
P.~Holmes and J.~Guckenheimer.
\newblock \emph{Nonlinear oscillations, dynamical systems, and bifurcations of
  vector fields}, volume~42 of \emph{Applied Mathematical Sciences}.
\newblock Springer-Verlag, Berlin, Heidelberg, 1983.

\bibitem[Holmes et~al.(2012)Holmes, Lumley, Berkooz, and
  Rowley]{Holmes2012book}
P.~Holmes, J.~L. Lumley, G.~Berkooz, and C.~W. Rowley.
\newblock \emph{Turbulence, Coherent Structures, Dynamical Systems and
  Symmetry}.
\newblock Cambridge University Press, Cambridge, 2nd paperback edition, 2012.

\bibitem[Hopf(1950)]{hopf50}
E.~Hopf.
\newblock The partial differential equation $u_t + uu_x = \mu u_{xx}$.
\newblock \emph{Comm. Pure App. Math.}, 3:\penalty0 201--230, 1950.

\bibitem[Kaptanoglu et~al.(2021{\natexlab{a}})Kaptanoglu, Callaham, Hansen,
  Aravkin, and Brunton]{kaptanoglu2021promoting}
A.~A. Kaptanoglu, J.~L. Callaham, C.~J. Hansen, A.~Aravkin, and S.~L. Brunton.
\newblock Promoting global stability in data-driven models of quadratic
  nonlinear dynamics.
\newblock \emph{arXiv preprint arXiv:2105.01843}, 2021{\natexlab{a}}.

\bibitem[Kaptanoglu et~al.(2021{\natexlab{b}})Kaptanoglu, de~Silva, Fasel,
  Kaheman, Callaham, Delahunt, Champion, Loiseau, Kutz, and
  Brunton]{kaptanoglu2021pysindy}
A.~A. Kaptanoglu, B.~M. de~Silva, U.~Fasel, K.~Kaheman, J.~L. Callaham, C.~B.
  Delahunt, K.~Champion, J.-C. Loiseau, J.~N. Kutz, and S.~L. Brunton.
\newblock Pysindy: A comprehensive python package for robust sparse system
  identification.
\newblock \emph{arXiv preprint arXiv:2111.08481}, 2021{\natexlab{b}}.

\bibitem[Kaptanoglu et~al.(2021{\natexlab{c}})Kaptanoglu, Morgan, Hansen, and
  Brunton]{kaptanoglu2020physics}
A.~A. Kaptanoglu, K.~D. Morgan, C.~J. Hansen, and S.~L. Brunton.
\newblock Physics-constrained, low-dimensional models for mhd: First-principles
  and data-driven approaches.
\newblock \emph{Physical Review E}, 104\penalty0 (015206), 2021{\natexlab{c}}.

\bibitem[Karniadakis et~al.(2021)Karniadakis, Kevrekidis, Lu, Perdikaris, Wang,
  and Yang]{karniadakis2021physics}
G.~E. Karniadakis, I.~G. Kevrekidis, L.~Lu, P.~Perdikaris, S.~Wang, and
  L.~Yang.
\newblock Physics-informed machine learning.
\newblock \emph{Nature Reviews Physics}, 3\penalty0 (6):\penalty0 422--440,
  2021.

\bibitem[Klus et~al.(2018)Klus, N{\"u}ske, Koltai, Wu, Kevrekidis, Sch{\"u}tte,
  and No{\'e}]{klus2017data}
S.~Klus, F.~N{\"u}ske, P.~Koltai, H.~Wu, I.~Kevrekidis, C.~Sch{\"u}tte, and
  F.~No{\'e}.
\newblock Data-driven model reduction and transfer operator approximation.
\newblock \emph{Journal of Nonlinear Science}, pages 1--26, 2018.

\bibitem[Kochkov et~al.(2021)Kochkov, Smith, Alieva, Wang, Brenner, and
  Hoyer]{kochkov2021machine}
D.~Kochkov, J.~A. Smith, A.~Alieva, Q.~Wang, M.~P. Brenner, and S.~Hoyer.
\newblock Machine learning accelerated computational fluid dynamics.
\newblock \emph{arXiv preprint arXiv:2102.01010}, 2021.

\bibitem[Koopman(1931)]{Koopman1931pnas}
B.~O. Koopman.
\newblock Hamiltonian systems and transformation in {H}ilbert space.
\newblock \emph{Proceedings of the National Academy of Sciences}, 17\penalty0
  (5):\penalty0 315--318, 1931.
\newblock URL \url{http://www.pnas.org/content/17/5/315.short}.

\bibitem[Kovachki et~al.(2021)Kovachki, Li, Liu, Azizzadenesheli, Bhattacharya,
  Stuart, and Anandkumar]{kovachki2021neural}
N.~Kovachki, Z.~Li, B.~Liu, K.~Azizzadenesheli, K.~Bhattacharya, A.~Stuart, and
  A.~Anandkumar.
\newblock Neural operator: Learning maps between function spaces.
\newblock \emph{arXiv preprint arXiv:2108.08481}, 2021.

\bibitem[Kutz et~al.(2016)Kutz, Brunton, Brunton, and Proctor]{Kutz2016book}
J.~N. Kutz, S.~L. Brunton, B.~W. Brunton, and J.~L. Proctor.
\newblock \emph{Dynamic Mode Decomposition: Data-Driven Modeling of Complex
  Systems}.
\newblock SIAM, 2016.

\bibitem[Kutz et~al.(2018)Kutz, Proctor, and Brunton]{kutzPDE}
J.~N. Kutz, J.~L. Proctor, and S.~L. Brunton.
\newblock Applied {Koopman} theory for partial differential equations and
  data-driven modeling of spatio-temporal systems.
\newblock \emph{Complexity}, 2018\penalty0 (6010634):\penalty0 1--16, 2018.

\bibitem[Lee and Carlberg(2020)]{lee2020model}
K.~Lee and K.~T. Carlberg.
\newblock Model reduction of dynamical systems on nonlinear manifolds using
  deep convolutional autoencoders.
\newblock \emph{Journal of Computational Physics}, 404:\penalty0 108973, 2020.

\bibitem[Li et~al.(2017)Li, Dietrich, Bollt, and Kevrekidis]{Li2017chaos}
Q.~Li, F.~Dietrich, E.~M. Bollt, and I.~G. Kevrekidis.
\newblock Extended dynamic mode decomposition with dictionary learning: A
  data-driven adaptive spectral decomposition of the {K}oopman operator.
\newblock \emph{Chaos: An Interdisciplinary Journal of Nonlinear Science},
  27\penalty0 (10):\penalty0 103111, 2017.

\bibitem[Li et~al.(2020{\natexlab{a}})Li, Kovachki, Azizzadenesheli, Liu,
  Bhattacharya, Stuart, and Anandkumar]{li2020fourier}
Z.~Li, N.~Kovachki, K.~Azizzadenesheli, B.~Liu, K.~Bhattacharya, A.~Stuart, and
  A.~Anandkumar.
\newblock Fourier neural operator for parametric partial differential
  equations.
\newblock \emph{arXiv preprint arXiv:2010.08895}, 2020{\natexlab{a}}.

\bibitem[Li et~al.(2020{\natexlab{b}})Li, Kovachki, Azizzadenesheli, Liu,
  Bhattacharya, Stuart, and Anandkumar]{li2020multipole}
Z.~Li, N.~Kovachki, K.~Azizzadenesheli, B.~Liu, K.~Bhattacharya, A.~Stuart, and
  A.~Anandkumar.
\newblock Multipole graph neural operator for parametric partial differential
  equations.
\newblock \emph{arXiv preprint arXiv:2006.09535}, 2020{\natexlab{b}}.

\bibitem[Li et~al.(2020{\natexlab{c}})Li, Kovachki, Azizzadenesheli, Liu,
  Bhattacharya, Stuart, and Anandkumar]{li2020neural}
Z.~Li, N.~Kovachki, K.~Azizzadenesheli, B.~Liu, K.~Bhattacharya, A.~Stuart, and
  A.~Anandkumar.
\newblock Neural operator: Graph kernel network for partial differential
  equations.
\newblock \emph{arXiv preprint arXiv:2003.03485}, 2020{\natexlab{c}}.

\bibitem[Li et~al.(2022)Li, Meidani, and Farimani]{li2022transformer}
Z.~Li, K.~Meidani, and A.~B. Farimani.
\newblock Transformer for partial differential equations' operator learning.
\newblock \emph{arXiv preprint arXiv:2205.13671}, 2022.

\bibitem[Ling et~al.(2016)Ling, Kurzawski, and Templeton]{Ling2016jfm}
J.~Ling, A.~Kurzawski, and J.~Templeton.
\newblock Reynolds averaged turbulence modelling using deep neural networks
  with embedded invariance.
\newblock \emph{Journal of Fluid Mechanics}, 807:\penalty0 155--166, 2016.

\bibitem[Loiseau(2020)]{Loiseau2020tcfd}
J.-C. Loiseau.
\newblock Data-driven modeling of the chaotic thermal convection in an annular
  thermosyphon.
\newblock \emph{Theoretical and Computational Fluid Dynamics}, 34, 2020.

\bibitem[Loiseau and Brunton(2018)]{Loiseau2017jfm}
J.-C. Loiseau and S.~L. Brunton.
\newblock Constrained sparse {Galerkin} regression.
\newblock \emph{Journal of Fluid Mechanics}, 838:\penalty0 42--67, 2018.

\bibitem[Loiseau et~al.(2018)Loiseau, Noack, and Brunton]{Loiseau2018jfm}
J.-C. Loiseau, B.~R. Noack, and S.~L. Brunton.
\newblock Sparse reduced-order modeling: sensor-based dynamics to full-state
  estimation.
\newblock \emph{Journal of Fluid Mechanics}, 844:\penalty0 459--490, 2018.

\bibitem[Long et~al.(2018)Long, Lu, Ma, and Dong]{long2018pde}
Z.~Long, Y.~Lu, X.~Ma, and B.~Dong.
\newblock Pde-net: Learning pdes from data.
\newblock In \emph{International Conference on Machine Learning}, pages
  3208--3216. PMLR, 2018.

\bibitem[Long et~al.(2019)Long, Lu, and Dong]{long2019pde}
Z.~Long, Y.~Lu, and B.~Dong.
\newblock Pde-net 2.0: Learning pdes from data with a numeric-symbolic hybrid
  deep network.
\newblock \emph{Journal of Computational Physics}, 399:\penalty0 108925, 2019.

\bibitem[Lu et~al.(2019)Lu, Jin, and Karniadakis]{lu2019deeponet}
L.~Lu, P.~Jin, and G.~E. Karniadakis.
\newblock Deeponet: Learning nonlinear operators for identifying differential
  equations based on the universal approximation theorem of operators.
\newblock \emph{arXiv preprint arXiv:1910.03193}, 2019.

\bibitem[Lu et~al.(2021)Lu, Jin, Pang, Zhang, and Karniadakis]{lu2021learning}
L.~Lu, P.~Jin, G.~Pang, Z.~Zhang, and G.~E. Karniadakis.
\newblock Learning nonlinear operators via deeponet based on the universal
  approximation theorem of operators.
\newblock \emph{Nature Machine Intelligence}, 3\penalty0 (3):\penalty0
  218--229, 2021.

\bibitem[Lumley(1970)]{Lumley:1970}
J.~Lumley.
\newblock Toward a turbulent constitutive relation.
\newblock \emph{Journal of Fluid Mechanics}, 41\penalty0 (02):\penalty0
  413--434, 1970.

\bibitem[Lusch et~al.(2018)Lusch, Kutz, and Brunton]{lusch2018deep}
B.~Lusch, J.~N. Kutz, and S.~L. Brunton.
\newblock Deep learning for universal linear embeddings of nonlinear dynamics.
\newblock \emph{Nature communications}, 9\penalty0 (1):\penalty0 4950, 2018.

\bibitem[Mardt et~al.(2018)Mardt, Pasquali, Wu, and No{\'e}]{Mardt2017arxiv}
A.~Mardt, L.~Pasquali, H.~Wu, and F.~No{\'e}.
\newblock {VAMP}nets: Deep learning of molecular kinetics.
\newblock \emph{Nature Communications}, 9\penalty0 (5), 2018.

\bibitem[Marsden and Ratiu(1999)]{Marsden:MS}
J.~E. Marsden and T.~S. Ratiu.
\newblock \emph{Introduction to mechanics and symmetry}.
\newblock Springer-Verlag, 2nd edition, 1999.

\bibitem[Messenger and Bortz(2021{\natexlab{a}})]{messenger2021bweak}
D.~A. Messenger and D.~M. Bortz.
\newblock Weak sindy: Galerkin-based data-driven model selection.
\newblock \emph{Multiscale Modeling \& Simulation}, 19\penalty0 (3):\penalty0
  1474--1497, 2021{\natexlab{a}}.

\bibitem[Messenger and Bortz(2021{\natexlab{b}})]{messenger2021weak}
D.~A. Messenger and D.~M. Bortz.
\newblock Weak sindy for partial differential equations.
\newblock \emph{Journal of Computational Physics}, page 110525,
  2021{\natexlab{b}}.

\bibitem[Mezi{\'c}(2005)]{Mezic2005nd}
I.~Mezi{\'c}.
\newblock Spectral properties of dynamical systems, model reduction and
  decompositions.
\newblock \emph{Nonlinear Dynamics}, 41\penalty0 (1-3):\penalty0 309--325,
  2005.

\bibitem[Mezi\'c(2013)]{Mezic2013arfm}
I.~Mezi\'c.
\newblock Analysis of fluid flows via spectral properties of the {K}oopman
  operator.
\newblock \emph{Ann. Rev. Fluid Mech.}, 45:\penalty0 357--378, 2013.

\bibitem[Mezi{\'c} and Banaszuk(2004)]{Mezic2004}
I.~Mezi{\'c} and A.~Banaszuk.
\newblock Comparison of systems with complex behavior.
\newblock \emph{Physica D: Nonlinear Phenomena}, 197\penalty0 (1--2):\penalty0
  101 -- 133, 2004.
\newblock ISSN 0167-2789.
\newblock \doi{http://dx.doi.org/10.1016/j.physd.2004.06.015}.
\newblock URL
  \url{http://www.sciencedirect.com/science/article/pii/S0167278904002507}.

\bibitem[Mollenhauer et~al.(2022)Mollenhauer, M{\"u}cke, and
  Sullivan]{mollenhauer2022learning}
M.~Mollenhauer, N.~M{\"u}cke, and T.~Sullivan.
\newblock Learning linear operators: Infinite-dimensional regression as a
  well-behaved non-compact inverse problem.
\newblock \emph{arXiv preprint arXiv:2211.08875}, 2022.

\bibitem[No{\'e} and Nuske(2013)]{noe2013variational}
F.~No{\'e} and F.~Nuske.
\newblock A variational approach to modeling slow processes in stochastic
  dynamical systems.
\newblock \emph{Multiscale Modeling \& Simulation}, 11\penalty0 (2):\penalty0
  635--655, 2013.

\bibitem[N\"{u}ske et~al.(2014)N\"{u}ske, Keller, P{\'e}rez-Hern{\'a}ndez, Mey,
  and No{\'e}]{nuske2014variational}
F.~N\"{u}ske, B.~G. Keller, G.~P{\'e}rez-Hern{\'a}ndez, A.~S. Mey, and
  F.~No{\'e}.
\newblock Variational approach to molecular kinetics.
\newblock \emph{Journal of chemical theory and computation}, 10\penalty0
  (4):\penalty0 1739--1752, 2014.

\bibitem[Otto and Rowley(2017)]{Otto2017arxiv}
S.~E. Otto and C.~W. Rowley.
\newblock Linearly-recurrent autoencoder networks for learning dynamics.
\newblock \emph{arXiv preprint arXiv:1712.01378}, 2017.

\bibitem[Page and Kerswell(2018)]{page2018koopman}
J.~Page and R.~R. Kerswell.
\newblock Koopman analysis of burgers equation.
\newblock \emph{Physical Review Fluids}, 3\penalty0 (7):\penalty0 071901, 2018.

\bibitem[Pan et~al.(2023)Pan, Brunton, and Kutz]{pan2023neural}
S.~Pan, S.~L. Brunton, and J.~N. Kutz.
\newblock Neural implicit flow: a mesh-agnostic dimensionality reduction
  paradigm of spatio-temporal data.
\newblock \emph{Journal of Machine Learning Research}, 24\penalty0
  (41):\penalty0 1--60, 2023.

\bibitem[Pathak et~al.(2017)Pathak, Lu, Hunt, Girvan, and Ott]{pathak2017using}
J.~Pathak, Z.~Lu, B.~R. Hunt, M.~Girvan, and E.~Ott.
\newblock Using machine learning to replicate chaotic attractors and calculate
  lyapunov exponents from data.
\newblock \emph{Chaos: An Interdisciplinary Journal of Nonlinear Science},
  27\penalty0 (12):\penalty0 121102, 2017.

\bibitem[Pathak et~al.(2018)Pathak, Hunt, Girvan, Lu, and Ott]{pathak2018model}
J.~Pathak, B.~Hunt, M.~Girvan, Z.~Lu, and E.~Ott.
\newblock Model-free prediction of large spatiotemporally chaotic systems from
  data: a reservoir computing approach.
\newblock \emph{Physical review letters}, 120\penalty0 (2):\penalty0 024102,
  2018.

\bibitem[Peherstorfer and Willcox(2016)]{peherstorfer2016data}
B.~Peherstorfer and K.~Willcox.
\newblock Data-driven operator inference for nonintrusive projection-based
  model reduction.
\newblock \emph{Computer Methods in Applied Mechanics and Engineering},
  306:\penalty0 196--215, 2016.

\bibitem[Peherstorfer et~al.(2020)Peherstorfer, Drmac, and
  Gugercin]{peherstorfer2020stability}
B.~Peherstorfer, Z.~Drmac, and S.~Gugercin.
\newblock Stability of discrete empirical interpolation and gappy proper
  orthogonal decomposition with randomized and deterministic sampling points.
\newblock \emph{SIAM Journal on Scientific Computing}, 42\penalty0
  (5):\penalty0 A2837--A2864, 2020.

\bibitem[Pope(1975)]{pope1975more}
S.~Pope.
\newblock A more general effective-viscosity hypothesis.
\newblock \emph{Journal of Fluid Mechanics}, 72\penalty0 (2):\penalty0
  331--340, 1975.

\bibitem[Qian et~al.(2020)Qian, Kramer, Peherstorfer, and
  Willcox]{qian2020lift}
E.~Qian, B.~Kramer, B.~Peherstorfer, and K.~Willcox.
\newblock Lift \& learn: Physics-informed machine learning for large-scale
  nonlinear dynamical systems.
\newblock \emph{Physica D: Nonlinear Phenomena}, 406:\penalty0 132401, 2020.

\bibitem[Raissi et~al.(2019)Raissi, Perdikaris, and Karniadakis]{Raissi2019jcp}
M.~Raissi, P.~Perdikaris, and G.~Karniadakis.
\newblock Physics-informed neural networks: A deep learning framework for
  solving forward and inverse problems involving nonlinear partial differential
  equations.
\newblock \emph{Journal of Computational Physics}, 378:\penalty0 686--707,
  2019.

\bibitem[Reed and Simon(1980)]{reed1980}
M.~Reed and B.~Simon.
\newblock \emph{Methods of Modern Mathematical Physics. {{I}}}.
\newblock {Academic Press Inc. [Harcourt Brace Jovanovich Publishers]}, {New
  York}, second edition, 1980.

\bibitem[Reinbold et~al.(2020)Reinbold, Gurevich, and
  Grigoriev]{Reinbold2020pre}
P.~A. Reinbold, D.~R. Gurevich, and R.~O. Grigoriev.
\newblock Using noisy or incomplete data to discover models of spatiotemporal
  dynamics.
\newblock \emph{Physical Review E}, 101\penalty0 (1):\penalty0 010203, 2020.

\bibitem[Reinbold et~al.(2021)Reinbold, Kageorge, Schatz, and
  Grigoriev]{reinbold2021robust}
P.~A. Reinbold, L.~M. Kageorge, M.~F. Schatz, and R.~O. Grigoriev.
\newblock Robust learning from noisy, incomplete, high-dimensional experimental
  data via physically constrained symbolic regression.
\newblock \emph{Nature communications}, 12\penalty0 (1):\penalty0 1--8, 2021.

\bibitem[Rowley and Marsden(2000)]{mars_rowl}
C.~W. Rowley and J.~E. Marsden.
\newblock Reconstruction equations and the {K}arhunen--{L}o{\`e}ve expansion
  for systems with symmetry.
\newblock \emph{Physica D: Nonlinear Phenomena}, 142\penalty0 (1):\penalty0
  1--19, 2000.

\bibitem[Rowley et~al.(2009)Rowley, Mezi\'c, Bagheri, Schlatter, and
  Henningson]{Rowley2009jfm}
C.~W. Rowley, I.~Mezi\'c, S.~Bagheri, P.~Schlatter, and D.~Henningson.
\newblock Spectral analysis of nonlinear flows.
\newblock \emph{J.\ Fluid Mech.}, 645:\penalty0 115--127, 2009.

\bibitem[Rudy et~al.(2017)Rudy, Brunton, Proctor, and Kutz]{Rudy2017sciadv}
S.~H. Rudy, S.~L. Brunton, J.~L. Proctor, and J.~N. Kutz.
\newblock Data-driven discovery of partial differential equations.
\newblock \emph{Science Advances}, 3\penalty0 (e1602614), 2017.

\bibitem[Sanchez-Gonzalez et~al.(2020)Sanchez-Gonzalez, Godwin, Pfaff, Ying,
  Leskovec, and Battaglia]{sanchez2020learning}
A.~Sanchez-Gonzalez, J.~Godwin, T.~Pfaff, R.~Ying, J.~Leskovec, and
  P.~Battaglia.
\newblock Learning to simulate complex physics with graph networks.
\newblock In \emph{International Conference on Machine Learning}, pages
  8459--8468. PMLR, 2020.

\bibitem[Schaeffer(2017)]{Schaeffer2017prsa}
H.~Schaeffer.
\newblock Learning partial differential equations via data discovery and sparse
  optimization.
\newblock In \emph{Proc. R. Soc. A}, volume 473, page 20160446. The Royal
  Society, 2017.

\bibitem[Schaeffer and McCalla(2017)]{Schaeffer2017pre}
H.~Schaeffer and S.~G. McCalla.
\newblock Sparse model selection via integral terms.
\newblock \emph{Physical Review E}, 96\penalty0 (2):\penalty0 023302, 2017.

\bibitem[Schaeffer et~al.(2013)Schaeffer, Caflisch, Hauck, and
  Osher]{Schaeffer2013pnas}
H.~Schaeffer, R.~Caflisch, C.~D. Hauck, and S.~Osher.
\newblock Sparse dynamics for partial differential equations.
\newblock \emph{Proceedings of the National Academy of Sciences USA},
  110\penalty0 (17):\penalty0 6634--6639, 2013.

\bibitem[Schmelzer et~al.(2020)Schmelzer, Dwight, and
  Cinnella]{schmelzer2020discovery}
M.~Schmelzer, R.~P. Dwight, and P.~Cinnella.
\newblock Discovery of algebraic reynolds-stress models using sparse symbolic
  regression.
\newblock \emph{Flow, Turbulence and Combustion}, 104\penalty0 (2):\penalty0
  579--603, 2020.

\bibitem[Schmid(2010)]{Schmid2010jfm}
P.~J. Schmid.
\newblock Dynamic mode decomposition for numerical and experimental data.
\newblock \emph{J.\ Fluid. Mech}, 656:\penalty0 5--28, 2010.

\bibitem[Schmidt and Lipson(2009)]{Schmidt2009science}
M.~Schmidt and H.~Lipson.
\newblock Distilling free-form natural laws from experimental data.
\newblock \emph{Science}, 324\penalty0 (5923):\penalty0 81--85, 2009.

\bibitem[Sharma and McKeon(2013)]{sharma2013coherent}
A.~Sharma and B.~McKeon.
\newblock On coherent structure in wall turbulence.
\newblock \emph{Journal of Fluid Mechanics}, 728:\penalty0 196--238, 2013.

\bibitem[Stevens and Colonius(2020{\natexlab{a}})]{stevens2020enhancement}
B.~Stevens and T.~Colonius.
\newblock Enhancement of shock-capturing methods via machine learning.
\newblock \emph{Theoretical and Computational Fluid Dynamics}, 34:\penalty0
  483--496, 2020{\natexlab{a}}.

\bibitem[Stevens and Colonius(2020{\natexlab{b}})]{stevens2020finitenet}
B.~Stevens and T.~Colonius.
\newblock Finitenet: A fully convolutional lstm network architecture for
  time-dependent partial differential equations.
\newblock \emph{arXiv preprint arXiv:2002.03014}, 2020{\natexlab{b}}.

\bibitem[Supekar et~al.(2023)Supekar, Song, Hastewell, Choi, Mietke, and
  Dunkel]{supekar2023learning}
R.~Supekar, B.~Song, A.~Hastewell, G.~P. Choi, A.~Mietke, and J.~Dunkel.
\newblock Learning hydrodynamic equations for active matter from particle
  simulations and experiments.
\newblock \emph{Proceedings of the National Academy of Sciences}, 120\penalty0
  (7):\penalty0 e2206994120, 2023.

\bibitem[Suri et~al.(2020)Suri, Kageorge, Grigoriev, and
  Schatz]{suri2020capturing}
B.~Suri, L.~Kageorge, R.~O. Grigoriev, and M.~F. Schatz.
\newblock Capturing turbulent dynamics and statistics in experiments with
  unstable periodic orbits.
\newblock \emph{Physical Review Letters}, 125\penalty0 (6):\penalty0 064501,
  2020.

\bibitem[Taira et~al.(2017)Taira, Brunton, Dawson, Rowley, Colonius, McKeon,
  Schmidt, Gordeyev, Theofilis, and Ukeiley]{Taira2017aiaa}
K.~Taira, S.~L. Brunton, S.~Dawson, C.~W. Rowley, T.~Colonius, B.~J. McKeon,
  O.~T. Schmidt, S.~Gordeyev, V.~Theofilis, and L.~S. Ukeiley.
\newblock Modal analysis of fluid flows: An overview.
\newblock \emph{AIAA Journal}, 55\penalty0 (12):\penalty0 4013--4041, 2017.

\bibitem[Taira et~al.(2020)Taira, Hemati, Brunton, Sun, Duraisamy, Bagheri,
  Dawson, and Yeh]{Taira2020aiaaj}
K.~Taira, M.~S. Hemati, S.~L. Brunton, Y.~Sun, K.~Duraisamy, S.~Bagheri,
  S.~Dawson, and C.-A. Yeh.
\newblock Modal analysis of fluid flows: Applications and outlook.
\newblock \emph{AIAA Journal}, 58\penalty0 (3):\penalty0 998--1022, 2020.

\bibitem[Takamoto et~al.(2022)Takamoto, Praditia, Leiteritz, MacKinlay,
  Alesiani, Pfl{\"u}ger, and Niepert]{takamoto2022pdebench}
M.~Takamoto, T.~Praditia, R.~Leiteritz, D.~MacKinlay, F.~Alesiani,
  D.~Pfl{\"u}ger, and M.~Niepert.
\newblock Pdebench: An extensive benchmark for scientific machine learning.
\newblock \emph{arXiv preprint arXiv:2210.07182}, 2022.

\bibitem[Takeishi et~al.(2017)Takeishi, Kawahara, and Yairi]{Takeishi2017nips}
N.~Takeishi, Y.~Kawahara, and T.~Yairi.
\newblock Learning {K}oopman invariant subspaces for dynamic mode
  decomposition.
\newblock In \emph{Advances in Neural Information Processing Systems}, pages
  1130--1140, 2017.

\bibitem[Vinuesa and Brunton(2022)]{vinuesa2022enhancing}
R.~Vinuesa and S.~L. Brunton.
\newblock Enhancing computational fluid dynamics with machine learning.
\newblock \emph{Nature Computational Science}, 2\penalty0 (6):\penalty0
  358--366, 2022.

\bibitem[Wang et~al.(2020{\natexlab{a}})Wang, Kashinath, Mustafa, Albert, and
  Yu]{wang2020towards}
R.~Wang, K.~Kashinath, M.~Mustafa, A.~Albert, and R.~Yu.
\newblock Towards physics-informed deep learning for turbulent flow prediction.
\newblock In \emph{Proceedings of the 26th ACM SIGKDD International Conference
  on Knowledge Discovery \& Data Mining}, pages 1457--1466, 2020{\natexlab{a}}.

\bibitem[Wang et~al.(2020{\natexlab{b}})Wang, Walters, and
  Yu]{wang2020incorporating}
R.~Wang, R.~Walters, and R.~Yu.
\newblock Incorporating symmetry into deep dynamics models for improved
  generalization.
\newblock \emph{arXiv preprint arXiv:2002.03061}, 2020{\natexlab{b}}.

\bibitem[Wehmeyer and No{\'e}(2018)]{Wehmeyer2017arxiv}
C.~Wehmeyer and F.~No{\'e}.
\newblock Time-lagged autoencoders: Deep learning of slow collective variables
  for molecular kinetics.
\newblock \emph{The Journal of Chemical Physics}, 148\penalty0 (241703), 2018.

\bibitem[Williams et~al.(2015{\natexlab{a}})Williams, Kevrekidis, and
  Rowley]{Williams2015jnls}
M.~O. Williams, I.~G. Kevrekidis, and C.~W. Rowley.
\newblock A data-driven approximation of the {K}oopman operator: extending
  dynamic mode decomposition.
\newblock \emph{Journal of Nonlinear Science}, 6:\penalty0 1307--1346,
  2015{\natexlab{a}}.

\bibitem[Williams et~al.(2015{\natexlab{b}})Williams, Rowley, and
  Kevrekidis]{Williams2014arxivA}
M.~O. Williams, C.~W. Rowley, and I.~G. Kevrekidis.
\newblock A kernel approach to data-driven {K}oopman spectral analysis.
\newblock \emph{Journal of Computational Dynamics}, 2:\penalty0 247,
  2015{\natexlab{b}}.

\bibitem[Yeung et~al.(2017)Yeung, Kundu, and Hodas]{Yeung2017arxiv}
E.~Yeung, S.~Kundu, and N.~Hodas.
\newblock Learning deep neural network representations for {K}oopman operators
  of nonlinear dynamical systems.
\newblock \emph{arXiv preprint arXiv:1708.06850}, 2017.

\bibitem[Yousif et~al.(2022)Yousif, Zhang, Yu, Vinuesa, and
  Lim]{yousif2022transformer}
M.~Z. Yousif, M.~Zhang, L.~Yu, R.~Vinuesa, and H.~Lim.
\newblock A transformer-based synthetic-inflow generator for
  spatially-developing turbulent boundary layers.
\newblock \emph{arXiv preprint arXiv:2206.01618}, 2022.

\bibitem[Zanna and Bolton(2020)]{zanna2020data}
L.~Zanna and T.~Bolton.
\newblock Data-driven equation discovery of ocean mesoscale closures.
\newblock \emph{Geophysical Research Letters}, 47\penalty0 (17):\penalty0
  e2020GL088376, 2020.

\end{thebibliography}

\end{spacing}
\end{document}